\documentclass[10pt,twocolumn,letterpaper]{article}

\usepackage{3dv}
\usepackage{times}
\usepackage{epsfig}
\usepackage{graphicx}
\usepackage{amsmath}
\usepackage{amssymb}
  
\usepackage[pagebackref=true,breaklinks=true,colorlinks,bookmarks=false]{hyperref}

\threedvfinalcopy 

\def\threedvPaperID{74} 

\usepackage{graphicx}
\usepackage{amsmath}
\usepackage{amssymb}
\usepackage{booktabs}

\usepackage{caption}
\usepackage{multirow}
\usepackage{mathptmx}

\usepackage{wrapfig}

\usepackage{times}
\usepackage{epsfig}
\usepackage{graphicx}
\usepackage{amsmath}
\usepackage{amssymb}
\usepackage{makecell}
\usepackage{caption}
\usepackage{wrapfig}
\usepackage{booktabs}

\usepackage{pifont}
\newcommand{\cmark}{\ding{51}}%
\newcommand{\xmark}{\ding{55}}%

\newcommand\cnum[1]{\raisebox{.5pt}{\textcircled{\raisebox{-0.9pt}{#1}}}}

\DeclareMathOperator*{\argmin}{arg\,min}

\newcommand\bigs[1]{\scalebox{1.2}{$#1$}}

\usepackage[square, comma, sort&compress, numbers]{natbib}

\usepackage[capitalize]{cleveref}
\crefname{section}{Sec.}{Secs.}
\Crefname{section}{Section}{Sections}
\Crefname{table}{Table}{Tables}
\crefname{table}{Tab.}{Tabs.}

\usepackage{overpic}
\usepackage{enumitem} 
\usepackage{overpic} 
\usepackage{color}

\definecolor{turquoise}{cmyk}{0.65,0,0.1,0.3}
\definecolor{purple}{rgb}{0.65,0,0.65}
\definecolor{dark_green}{rgb}{0, 0.5, 0}
\definecolor{orange}{rgb}{0.8, 0.6, 0.2}
\definecolor{red}{rgb}{0.8, 0.2, 0.2}
\definecolor{darkred}{rgb}{0.6, 0.1, 0.05}
\definecolor{blueish}{rgb}{0.0, 0.3, .6}
\definecolor{light_gray}{rgb}{0.7, 0.7, .7}
\definecolor{pink}{rgb}{1, 0, 1}
\definecolor{greyblue}{rgb}{0.25, 0.25, 1}

\usepackage{blindtext}

\renewcommand{\paragraph}[1]{\vspace{1em}\noindent\textbf{#1}.}

\begin{document}
	  
	\title{HVTR: Hybrid Volumetric-Textural Rendering for Human Avatars}

\author{Tao Hu$^{1}\thanks{Work partly done during TH's internship at Tsinghua University.}$, ~Tao Yu$^{2}$, ~Zerong Zheng$^{2}$, ~He Zhang$^{2}$, ~Yebin Liu$^{2}\thanks{Corresponding author.}$, ~Matthias Zwicker$^{1}$ \\
	\vspace{0.12in}
	$^1$University of Maryland, College Park  ~~$^2$Tsinghua University	
}



	\maketitle
	    

\begin{abstract}
    We propose a novel neural rendering pipeline, Hybrid Volumetric-Textural Rendering (HVTR), which synthesizes virtual human avatars from arbitrary poses efficiently and at high quality. First, we learn to encode articulated human motions on a dense UV manifold of the human body surface. To handle complicated motions (e.g., self-occlusions), we then leverage the encoded information on the UV manifold to construct a 3D volumetric representation based on a dynamic pose-conditioned neural radiance field. While this allows us to represent 3D geometry with changing topology, volumetric rendering is computationally heavy. Hence we employ only a rough volumetric representation using a pose-conditioned downsampled neural radiance field (PD-NeRF), which we can render efficiently at low resolutions. In addition, we learn 2D textural features that are fused with rendered volumetric features in image space. The key advantage of our approach is that we can then convert the fused features into a high-resolution, high-quality avatar by a fast GAN-based textural renderer. We demonstrate that hybrid rendering enables HVTR to handle complicated motions, render high-quality avatars under user-controlled poses/shapes and even loose clothing, and most importantly, be efficient at inference time. Our experimental results also demonstrate state-of-the-art quantitative results. More results are available at our project page: \url{https://www.cs.umd.edu/~taohu/hvtr/}

\end{abstract}
	\section{Introduction}
\label{sec:intro}
\vspace{-0.2pt}
Capturing and rendering realistic human appearance under varying poses and viewpoints is an important goal in computer vision and graphics. Recent neural rendering methods \cite{Tewari2020StateOT,neuralbody,neuralactor,anr,Weng2022HumanNeRFFR,Zhao2021HumanNeRFGN,Lombardi2021MixtureOV} have made great progress in generating realistic images of humans, which are simple yet effective compared with traditional graphics pipelines \cite{Borshukov2003UniversalCI,Carranza2003FreeviewpointVO,Xu2011VideobasedCC}. 

Given a training dataset of multiple synchronized RGB videos of a human, the goal is to build an animatable virtual avatar with pose-dependent geometry and appearance of the individual that can be driven by arbitrary poses from arbitrary viewpoints at inference time. We propose Hybrid Volumetric-Textural Rendering (HVTR). To represent the input to our system, including the pose and the rough body shape of an individual, we employ a skinned parameterized mesh (SMPL \cite{smpl}) fitted to the training videos. 
Our system is expected to handle the articulated structure of human bodies, various clothing styles, non-rigid motions, and self-occlusions, and be efficient at inference time. In the following, we will introduce how HVTR solves these challenges by proposing (1) effective pose encoding for better generalization, (2) rough yet effective volumetric representation to handle changing topology, and (3) hybrid rendering for efficient and high quality rendering.


\textit{Pose Encoding on a 2D Manifold}. The first challenge lies in encoding the input pose information so that it can be leveraged effectively by the rendering pipeline. Existing methods parameterize poses by global pose parameter conditioning \cite{Yang2018AnalyzingCL,Patel2020TailorNetPC,Ma2020LearningTD,Lhner2018DeepWrinklesAA}, 3D sparse points \cite{neuralbody}, or skinning weights \cite{neuralactor,Peng2021AnimatableNR,Chen2021AnimatableNR}. In contrast, we encode poses on a 2D UV manifold of the body mesh surface, and the dense representation enables us to utilize 2D convolutional networks to effectively encode pose features. In addition, we define a set of geometry and texture latents on the 2D manifold to capture local motion and appearance details for rendering. 

\textit{Rough Yet Effective Volumetric Representation}. Our input is a coarse SMPL mesh as used in \cite{smplpix,anr,egorend}, which cannot capture detailed pose- and clothing-dependent deformations. Inspired by the recent neural scene representations \cite{nerf,neuralactor,neuralbody,park2021nerfies,pumarola2020d}, we model articulated humans with an implicit volumetric representation by constructing a dynamic pose-conditioned neural radiance field. This volumetric representation has the built-in flexibility to handle changing geometry and topology. Different from NeRF \cite{nerf} for static scenes, we condition our proposed dynamic radiance field on our pose encoding defined on the UV manifold. This enables capturing pose- and view-dependent volumetric features. Constructing the radiance field is computationally heavy \cite{nerf,neuralactor,neuralbody}, however, hence we propose to learn only a rough volumetric representation by constructing a pose-conditioned downsampled neural radiance field (PD-NeRF). This allows us to balance the competing challenges of achieving computational complexity while still being able to effectively resolve self-occlusions. Yet learning PD-NeRF from low resolution images is challenging, and to address this, we propose an appropriate sampling scheme. We show that we can effectively train PD-NeRF from $45\times45$ images with as few as 7 sampled points along each query ray (see Fig.~\ref{fig:ab_all} and Tab.~\ref{tab:full_acc_time}).  HVTR is about 52$\times$ faster than Neural Body \cite{neuralbody} in inference (see Tab.~\ref{tab:full_acc_time}).


\textit{Hybrid Rendering}.
The final challenge is to render full resolution images by combining the downsampled PD-NeRF and our learned latents on the 2D UV manifold. To solve this, we rasterize the radiance field into multi-channel (not just RGB) volumetric features in image space by volume rendering. The rasterized volumetric features preserve both geometric and appearance details \cite{nerf}. In addition, we extract 2D textural features from our latents on the UV manifold for realistic image synthesis following the spirit of Deferred Neural Rendering (DNR) \cite{dnr,anr,egorend}. We fuse the 3D and 2D features by utilizing Attentional Feature Fusion (AFF \cite{dai21aff}), and finally use a 2D GAN-based \cite{gans} textural rendering network (TexRenderer) to decode and supersample them into realistic images. Though TexRenderer works in image space, it is able to incorporate the rasterized volumetric features for geometry-aware rendering.

The hybrid rendering brings several advantages. (1) We are able to handle self-occlusions by volume rendering. (2) We can generate high quality details using GAN and adversarial training. This enables us to handle uncertainties involved in modeling dynamic details, and is well-suited for enforcing realistic rendered images \cite{anr,Huang2020AdversarialTO,smplpix}. (3) We only need to learn rough geometries, which make our pipeline efficient in inference. (4) Benefiting from (1, 2, 3), we are able to handle loose clothing (Fig. \ref{fig:dress}). 

In summary, our contributions are: 
{(1)} We propose HVTR, a novel neural rendering pipeline, to generate human avatars from arbitrary skeleton/SMPL motions using a hybrid strategy. HVTR achieves SOTA performance, and is able to handle complicated motions, render high quality avatars even with loose clothing, generalize to novel poses, and support body shape control. Most importantly, it is efficient at inference time. 
{(2)} HVTR uses an effective scheme to encode pose information on the UV manifold of body surfaces, and leverages this to learn a pose-conditioned downsampled NeRF (PD-NeRF) from low resolution images. Our experiments show how the rendering quality is influenced by the PD-NeRF resolution, and that even low resolution volumetric representations can produce high quality outputs at a small computational cost. 
{(3)} HVTR shows how to construct PD-NeRF and extract 2D textural features all based on pose encoding on the UV manifold, and most importantly, how the two can be fused and incorporated for fast, high quality, and geometry-aware neural rendering.

\section{Related Work}
\label{sec:related}

\begin{figure*}[t]
	\begin{center}
		\includegraphics[width=\linewidth]{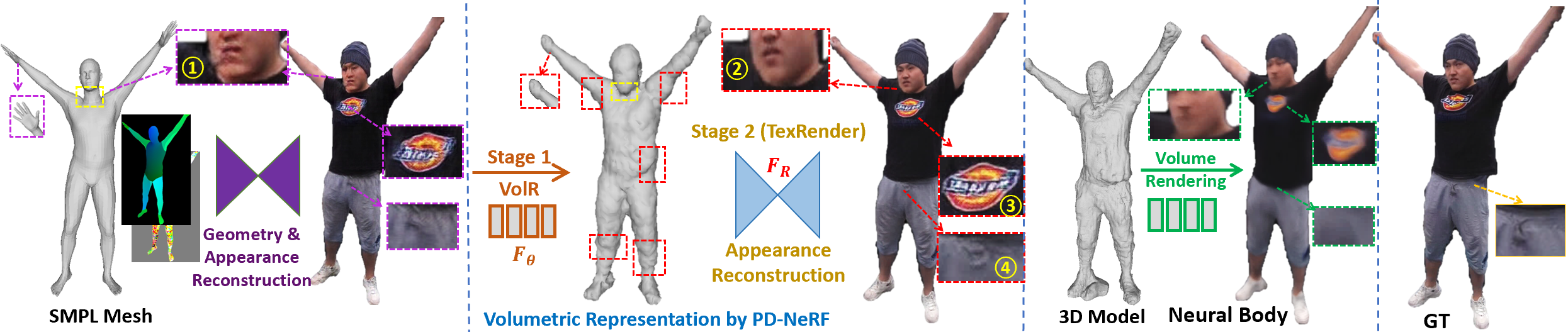}
	\end{center}
	\vspace{-0.18in}	
	\caption{We illustrate the differences between (left) GAN-based methods (DNR), (middle) our hybrid approach, and (right) NeRF methods (Neural Body \cite{neuralbody}). DNR\cite{dnr} and SMPLpix\cite{smplpix} are based on fixed mesh (SMPL\cite{smpl} or SMPL-X\cite{SMPLX}), and use a GAN for one-stage rendering
	without explicit geometry reconstruction. As a disadvantage, DNR needs to resolve geometric misalignments implicitly, which often leads to artifacts (see closeup \cnum{1} in the figure). Yet our method (middle) works in two stages by first learning a downsampled volumetric representation (by PD-NeRF), and then utilizing a GAN for appearance synthesis. Though only learned from low resolution images ($90\times90$ in this example),
    the rough volumetric representation still encodes more 3D pose-dependent features than SMPL, which enables us to handle self-occlusions (region \cnum{1} vs \cnum{2}), and preserve more details (\cnum{3}\cnum{4}) than DNR. In addition, our GAN-based renderer can generate high resolution wrinkles, whereas Neural Body cannot.
    Besides, our approach is about 52$\times$ faster than Neural Body in inference (Tab. \ref{tab:full_acc_time}).}
	\label{fig:cmp_vr_nr}
	\vspace{-0.16in}
\end{figure*}


\noindent \textbf{Neural Scene Representations.} Instead of explicitly modeling geometry, many neural rendering methods \cite{sitzmann2019deepvoxels,Sitzmann2019,nerf,dnr} propose to learn implicit representations of scenes, such as DeepVoxels \cite{sitzmann2019deepvoxels}, Neural Volumes \cite{Lombardi2019NeuralV}, SRNs \cite{Sitzmann2019}, or NeRF \cite{nerf}. In contrast to these static representations, we learn a dynamic radiance field on the UV manifold of human surfaces to model articulated human bodies.

\begin{table}[t]
	\begin{tabular}{llcc}
		\hline
		Animatable  Pipelines & \multirow{2}{*}{\begin{tabular}[c]{@{}l@{}}Rend- \\ erer \end{tabular}} & \multirow{2}{*}{\begin{tabular}[c]{@{}l@{}}Geom- \\ Recon\end{tabular}} & \multirow{2}{*}{\begin{tabular}[c]{@{}l@{}}Fast \\ Infer.\end{tabular}} \\ 
		&                                                                         &                                                                            &                                                                         \\ \hline
		2D : EDN \cite{edn}, vid2vid \cite{vid2vid} & GAN                                                             & {\xmark}                                                                 & {\cmark}                                                              \\ \hline
		\multirow{2}{*}{\begin{tabular}[c]{@{}l@{}}2D Plus : SMPLpix \cite{smplpix}, \\ DNR \cite{dnr}, ANR\cite{anr} \end{tabular}} & \multirow{2}{*}{GAN}                                                & \multirow{2}{*}{\xmark}                                                         & \multirow{2}{*}{\cmark}                                                      \\
		&                                                                         \\ \hline
		3D : NB\cite{neuralbody}, AniNeRF\cite{Peng2021AnimatableNR} & VolR                                                                      & \cmark                                                                          & \xmark                                                                       \\ \hline \hline
		3D : Ours                                                                   & Hybrid                                                                      & \cmark                                                                          & \cmark \\ \hline                                                                      
	\end{tabular}
	\vspace{-0.12in}
	\caption{A set of recent human synthesis approaches classified by feature representations (2D/3D) and renderers. VolR: volume rendering \cite{Kajiya1984RayTV}.} 
	\label{tab:methods}
	\vspace{-0.18in}
\end{table}

\noindent \textbf{Shape Representations of Human.} To capture detailed deformations of human bodies, most recent papers utilize implicit representations \cite{Mescheder2019OccupancyNL,Michalkiewicz2019DeepLS,Chen2019LearningIF,Park2019DeepSDFLC,Saito2019PIFuPI,Saito2020PIFuHDMP,Huang2020ARCHAR,Saito2021SCANimateWS,Mihajlovi2021LEAPLA,Wang2021MetaAvatarLA,Palafox2021NPMsNP,Tiwari2021NeuralGIFNG,Zheng2021PaMIRPM,Jeruzalski2020NASANA,Zheng2021DeepMultiCapPC, Xiu_2022_CVPR,He2021ARCHAC} or point clouds \cite{scale,pop} due to their topological flexibility. These methods aim at learning geometry from 3D datasets, whereas we synthesize human images of novel poses only from 2D RGB training images.


\noindent \textbf{Rendering Humans by Generative Adversarial Network (GAN).} Some existing approaches formulate the human rendering as neural image translation, i.e. they map the body pose given in the form of renderings of a skeleton~\cite{edn,SiaroSLS2017,Pumarola_2018_CVPR,KratzHPV2017,zhu2019progressive,vid2vid}, dense mesh~\cite{Liu2019,wang2018vid2vid,liu2020NeuralHumanRendering,feanet,Neverova2018,Grigorev2019CoordinateBasedTI} or joint position heatmaps~\cite{MaSJSTV2017,Aberman2019DeepVP,Ma18}, to real images. As summarized in Tab.~\ref{tab:methods}, EDN \cite{edn} and vid2vid \cite{vid2vid} utilize GAN \cite{gans} networks to learn a mapping from 2D poses to human images. To improve temporal stability and learn a better mapping, ``2D Plus'' methods \cite{dnr,smplpix,egorend} are conditioned on a coarse mesh (SMPL \cite{smpl}), and take as input additional geometry features, such as DNR (UV mapped features) \cite{dnr}, SMPLpix (+ depth map) \cite{smplpix}, and ANR (UV + normal map) \cite{anr}. A 2D ConvNet is often utilized for both shape completion and appearance synthesis in one stage \cite{dnr,smplpix}. However, \cite{dnr,smplpix,anr,stylepeople} do not reconstruct geometry explicitly and cannot handle self-occlusions effectively. In contrast, our rendering is conditioned on a learned 3D volumetric representation using a two-stage approach (see Fig.~\ref{fig:cmp_vr_nr}), which handles self-occlusion more effectively than \cite{dnr,smplpix,anr,stylepeople} that just take geometry priors (e.g., UV, depth or normal maps) as input (see Fig.~\ref{fig:ab_all}, \ref{fig:rgb5_zju}).

\noindent \textbf{Rendering Humans by Volume Rendering (VolR).} For stable view synthesis, recent papers \cite{neuralbody,neuralactor,Peng2021AnimatableNR,narf,anerf,Chen2021AnimatableNR, Xu2021HNeRFNR} propose to unify geometry reconstruction with view synthesis by volume rendering, which, however, is computationally heavy. In addition, the appearance synthesis (e.g., Neural Body \cite{neuralbody}) largely relies on the quality of geometry reconstruction, which is very challenging for dynamic humans, and imperfect geometry reconstruction will lead to blurry images (Fig.~\ref{fig:rgb5_zju}). Furthermore, most animatable NeRF methods \cite{neuralactor,Peng2021AnimatableNR,narf,anerf,Chen2021AnimatableNR} parameterize clothes with skinning by learning backward warping, and they cannot handle skirts. In contrast, our method only needs a forward skinning step, and can render skirts (Fig. \ref{fig:dress}) with GAN. A comparison of ours, GAN renderer, and VolR is shown in Fig.~\ref{fig:cmp_vr_nr}.

Ours is distinguished from most recent 3D-GAN \cite{Niemeyer2021GIRAFFERS,Zhou2021CIPS3DA3,Gu2021StyleNeRFAS,OrEl2021StyleSDFH3,Hong2021HeadNeRFAR,Chan2021EfficientG3} by rendering dynamic humans at high resolutions, and conditioning our rendering framework on the UV manifold of human body surfaces.

\begin{figure*}[t]
	\begin{center}
		\includegraphics[width=\linewidth]{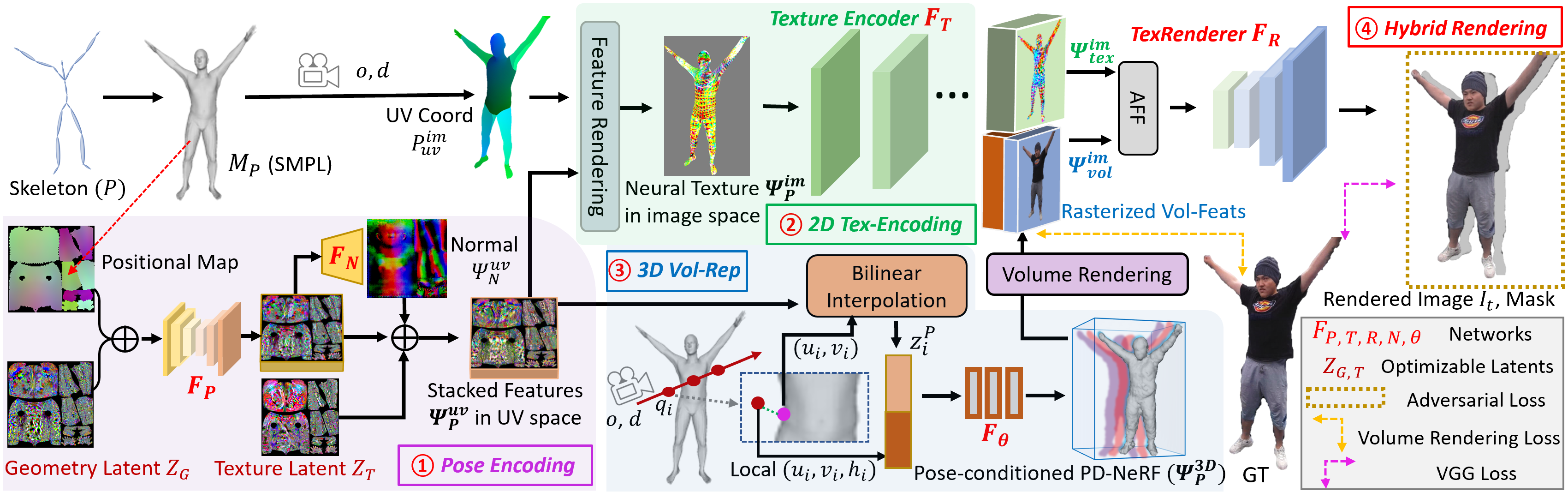}
	\end{center}
	\vspace{-0.16in}
	\caption{Pipeline overview. Given a coarse SMPL mesh \textbf{$M_P$} with pose \textit{{P}} and a target viewpoint (\textit{{o}}, \textit{{d}}), our system renders a target image $I_{ t}$ using four main components: \cnum{1} pose encoding, \cnum{2} 2D textural feature encoding, \cnum{3} 3D volumetric representation, and \cnum{4} hybrid rendering. \textbf{\cnum{1} Pose Encoding} in UV space: We record the 3D positions of the mesh on a UV positional map, and use a PoseNet $F_P$ to extract pose features. To enforce learning geometric features, we propose a NormalNet $F_N$ to predict the normals of SMPL mesh in UV space, and we propose a set of optimizable geometry latent $Z_G$ and texture style latent $Z_T$ to capture local motion and appearance details.  
	\textbf{\cnum{2} 2D Tex-Encoding}: A \textit{Feature Rendering} module renders the SMPL mesh into image features $\Psi^{im}_{P}$ by utilizing a rasterized UV coordinate map ($P^{im}_{uv}$). The image features are then encoded as 2D textural features $\Psi^{im}_{tex}$ by the \textit{Textural Encoder} ${F_T}$. \textbf{\cnum{3} 3D Vol-Rep}: To capture the rough geometry and address self-occlusion problems, we further learn a volumetric representation by constructing a pose-conditioned downsampled neural radiance field (PD-NeRF) to encode 3D pose-dependent features. \textbf{\cnum{4} Hybrid Rendering}: PD-NeRF is rasterized into image space $\Psi^{im}_{vol}$ by volume rendering, where 3D volumetric features are also preserved. Both the 2D $\Psi^{im}_{tex}$ and 3D features $\Psi^{im}_{vol}$ are pixel-aligned in image space, fused by Attentional Feature Fusion (AFF), and then converted into a realistic image $I_{ t}$ and a mask by TexRenderer ${F_R}$.  See the glossary term table in the Appendix.} 
	\label{fig:overview}
	\vspace{-0.16in}
\end{figure*}

\section{Method}
 Our goal is to render pose- and view-dependent avatars of an individual from an arbitrary pose {\textit{P}} and an arbitrary viewpoint (position {\textit{o}}, view direction {\textit{d}}):
$$ 
\vspace{-0.12pt}
I_{ t} = HTVR(\textit{P},~{\textit{o}},~{\textit{d}},~K) 
$$
where \textit{K} denotes the camera intrinsic parameters, and $I_{t} \in \mathbb{R}^{W \times H \times 3}$ is the target output image.

We first introduce our pose encoding method (Sec. \ref{sec:pose_encod}), and based on this we present how to extract 2D textural features (Sec. \ref{sec:tex_feat}) and 3D volumetric features (Sec. \ref{sec:vol_feat}). Finally, we describe how we fuse these features and synthesize the final RGB avatars (Sec. \ref{sec:tex_rendering}). Fig. \ref{fig:overview} shows an outline of the proposed framework. In the paper, we use $\Psi^{space}_{type}$ to denote the intermediate features in the pipeline.

\subsection{Pose Encoding}
\label{sec:pose_encod}
Given a skeleton pose {\textit{P}}, we first reconstruct a posed SMPL mesh \bigs{M_P} using the Linear Blend Skinning of SMPL \cite{smpl}. We construct a UV positional map (with a size of ${U\times U \times 3}$) by projecting each surface point on $\bigs{M_P}$ from 3D space to its UV manifold,  where each pixel describes the relative location of the point on the body surface manifold. With this, we define a geometry latent $Z_G \in \mathbb{R}^{U\times U\times C_g}$ to represent the intrinsic local geometry features, and a texture latent $Z_T \in \mathbb{R}^{U\times U\times C_t}$ to represent high-dimensional neural textures as used in \cite{dnr,anr,egorend}. Both latents are defined in UV space, and shared across different poses. Our geometry and texture latents have higher resolution than the compressed representation used in other works (e.g., latent vectors used in \cite{Palafox2021NPMsNP,Burov2021DynamicSF}), which enables us to capture local details, and the rendering pipeline can leverage them to infer local geometry and appearance changes. $Z_G$ and $Z_T$ are trainable tensors both with a size of $128\times 128 \times 16$.

The positional map and the geometry latent $Z_G$ are convolved by PoseNet \bigs{F_P} to obtain high-dimensional pose features. In addition, to enforce learning geometric features, we predict the normal $\Psi^{uv}_{N}$ of the posed mesh in UV space using NormalNet \bigs{F_N}. We then concatenate the geometric features and the texture latent $Z_T$ to obtain our pose-dependent features $\Psi^{uv}_{P}$.
 

Though the UV positional map used is similar to \cite{scale,pop}, ours is distinguished by learning pose-dependent features from 2D images instead of 3D data, and we have a normal estimation network to enforce geometric learning.


\subsection{2D Textural Feature Encoding}
\label{sec:tex_feat}
Given a viewpoint ({\textit{o}}, {\textit{d}}), we also render a UV coordinate map $P^{im}_{uv}$ to encode the shape and pose features of \bigs{M_P} in image space. This allows us to transform the pose-dependent features $\Psi^{uv}_{P}$ from UV space to image space $\Psi^{im}_{P}$ by utilizing a \textit{Feature Rendering} module. We further encode $\Psi^{im}_{P}$ as a high-dimensional textural feature using a Texture Encoder \bigs{F_T} implemented by a 2D ConvNet. 

\subsection{3D Volumetric Representation}
\label{sec:vol_feat}
Though existing methods achieve compelling view synthesis by just rendering with 2D textural features \cite{dnr,egorend,anr}, they cannot handle self-occlusions effectively since they do not reconstruct the geometry. We address this by learning a 3D volumetric representation using a pose-conditioned neural radiance field (PD-NeRF).

We include pose information to learn the volumetric representation by looking up the encoded pose features in $\Psi^{uv}_{P}$ corresponding to each 3D query point. To achieve this, we project each query point $q_i$ in the posed space of SMPL mesh \bigs{M_P} to a local point $\hat{q_i}=(u_i,v_i,h_i)$ in an  UV-plus-height space,
\begin{equation}
\vspace{-0.08pt}
(u_i,~v_i,~f_i) = \argmin_{u,~v,~f}{\|q_i-B_{u,v}(V_{[Tri(f)]})\|_2},
\end{equation}
where $f \in \{1,...,N_{Face}\}$ is the triangle index, $Tri(f)$ is the triangle (face), $V_{[Tri(f)]}$ are the three vertices of $Tri(f)$, $(u,v)$ are the barycentric coordinates of the face, and $B_{u,v}(.)$ is the barycentric interpolation function. The height $h_i$ is given by the signed distance of $q_i$ to the nearest face $Tri(f_i)$. 

With this, we sample the local feature $z^P_i$ from the encoded pose features $\Psi^{uv}_{P}$: $z^P_i = B_{u_i,v_i}(\Psi^{uv}_{P})$. Given a camera position \textit{{o}} and view direction \textit{{d}}, we predict the density {$\sigma$} and appearance features {$\xi$} of $q_i$ as
\begin{equation}
\vspace{-0.02in}
\bigs{F_{\theta}}: (\gamma(\hat{q_i}),~\gamma(d),~z^P_i) \rightarrow (\sigma, ~\xi),
\label{eq:nerf}
\end{equation}
where $\gamma$ is a positional encoder. $\xi \in {\mathbb{R}}^{\hbar}$ is a high-dimensional feature vector, where the first three channels are RGB colors. A key property of our approach is that $\bigs{F_{\theta}}$ is conditioned on high resolution encoded pose features $z^P_i$ instead of pose parameters {\textit{P}}.

\subsection{Hybrid Volumetric-Textural Rendering}
\label{sec:tex_rendering}
Though the radiance field PD-NeRF can be directly rendered into target images by volume rendering \cite{Kajiya1984RayTV}, this is computationally heavy. In addition, a direct deterministic regression using RGB images often leads to blurry results for stochastic clothing movements as stated in \cite{neuralactor}.

\noindent \textbf{Volumetric Rendering}. To address this, we use PD-NeRF to render downsampled images by a factor \textit{\textbf{s}} for efficient inference. We rasterize PD-NeRF into multi-channel volumetric features $\Psi^{im}_{vol} \in \mathbb{R}^{W_d \times H_d \times \hbar}$, and each pixel $\Re(r,~P)$ is predicted by \textit{N} consecutive samples $\{x_1,...,x_N\}$ along the corresponding ray \textit{r} through volume rendering \cite{Kajiya1984RayTV},
\begin{equation}
\vspace{-0.08in}
\Re(r,~P) = \sum_{n=1}^{N}(\prod_{i=1}^{n-1}e^{-\sigma_i \delta_i}) \cdot (1-e^{-\sigma_n \delta_n}) \cdot \xi_n,
\end{equation}
where $\delta_n = \|x_n - x_{n-1}\|_2$, and density and appearance features $\sigma_n$, $\xi_n$ of $x_n$ are predicted by Eq.~\ref{eq:nerf}. Note the first three channels of $\Psi^{im}_{vol}$ are RGB, which are supervised by downsampled ground truth images (see Fig. \ref{fig:overview}).
  
\noindent \textbf{Attentional Volumetric Textural Feature Fusion}. With both the 2D textural features $\Psi^{im}_{tex}$ and the 3D volumetric features $\Psi^{im}_{vol}$, the next step is to fuse them and leverage them for 2D image synthesis. This poses several challenges. First, $\Psi^{im}_{tex}$ is trained in 2D, which converges faster than $\Psi^{im}_{vol}$, since $\Psi^{im}_{vol}$ needs to regress a geometry by optimizing downsampled images, and NeRF training generally converges more slowly for dynamic scenes \cite{neuralbody}. Second, $\Psi^{im}_{tex}$ has higher dimensions (both resolution and channels) than $\Psi^{im}_{vol}$, because $\Psi^{im}_{vol}$ is learned from downsampled images with relatively weak supervision. Due to this, the system may tend to ignore the volumetric features of $\Psi^{im}_{vol}$. To solve this problem, we first use a ConvNet to downsample $\Psi^{im}_{tex}$ to the same size as $\Psi^{im}_{vol}$, and also extend the channels of $\Psi^{im}_{vol}$ to the same dimensionality as $\Psi^{im}_{tex}$ using a ConvNet.

Finally, we fuse the resized features by Attentional Feature Fusion (AFF \cite{dai21aff}): $\Psi^{im}_{vt} = AFF(\Psi^{im}_{vol}, \Psi^{im}_{tex})$, which has the same size as $\Psi^{im}_{vol}$. AFF is also learned, and we include it in $\bigs{F_R}$ in Fig.~\ref{fig:overview}. See \cite{dai21aff} for more details about AFF.
 
\noindent \textbf{Textural Rendering}. The TexRenderer net $\bigs{F_R}$ converts the fused features $\Psi^{im}_{vt}$ into the target avatar $I_{t}$ and a mask. $\bigs{F_R}$ has a similar architecture as Pix2PixHD \cite{pix2pixhd}.

\subsection{Optimization}
HVTR is trained end-to-end by optimizing networks $\bigs{F_{P,~T,~N,~R,~\theta}}$ and latent codes ${Z_{G,~T}}$. Given a ground truth image $I_{gt}$ and mask $M_{gt}$, downsampled ground truth image $I^D_{gt}$, and predicted image $I_{t}$ and mask $M_{t}$, we use the following loss functions:   

\noindent \textbf{Volume Rendering Loss}. We utilize ${L}_{vol}$ to supervise the training of volume rendering, which is applied on the first three channels of $\Psi^{im}_{vol}$, ${L}_{vol} = \|~I^D_{gt} ~-~ \Psi^{im}_{vol}[:3]~\|^2_2$.

\noindent \textbf{Normal Loss}. To enforce learning of geometric features by $\bigs{F_P}$, we employ a normal loss ${L}_{{norm}}$: ${L}_{{norm}} = \|~N^{uv}_{gt} - \Psi^{uv}_N~\|_1 $, where $N^{uv}_{gt}$ is the ground truth normal of mesh \bigs{M_P} projected into UV space.

\noindent \textbf{Feature Loss}. We use a feature loss \cite{Perceptual_Losses} to measure the differences between the activations on different layers of the pretrained VGG network \cite{vgg} of the generated image $I_{ t}$ and ground truth image $I_{gt}$,
\vspace{-0.08in}
\begin{equation}
\vspace{-0.08in}
{L}_{feat}=\sum \frac{1}{N^{j}}\left\|~g^{j}\left(I_{gt}\right) ~-~ g^{j}\left(I_{ t}\right)~\right\|_2,
\end{equation}
\noindent where $g^j$ is the activation and $N^j$ the number of elements of the $j$-th layer in the pretrained VGG network.

\noindent \textbf{Mask Loss}. The mask loss is ${L}_{mask} = \|~M_{gt} ~-~ M_{t}~\|_1$.

\noindent \textbf{Pixel Loss}. We also enforce an $\ell_1$ loss between the generated image and ground truth as ${L}_{pix} = \|~I_{gt} - I_{ t}~\|_1$.

\noindent \textbf{Adversarial Loss}. We leverage a multi-scale discriminator $D$ \cite{pix2pixhd}  as an adversarial loss ${L}_{adv}$. $D$ is conditioned on both the generated image and feature image $\Psi^{im}_{P}$.

\noindent \textbf{Face Identity Loss}. We use a pre-trained network to ensure that TexRenderer preserves the face identity on the cropped face of the generated and ground truth image,
\vspace{-0.08in}
\begin{equation}
\vspace{-0.08in}	
{L}_{{face}}= \|~ N_{ {face}}\left(I_{gt}\right) ~-~ N_{ {face}}\left(I_{ t}\right)\|_2,
\end{equation}
\noindent where $N_{face}$ is the pretrained SphereFaceNet \cite{Liu2017SphereFaceDH}.

\noindent \textbf{Total Loss}. 

${L}_{total} = \sum_{i\in \{vol,~norm,~feat,~mask,~pix,~adv,~face\}} \lambda_i {L}_i$.

The networks were trained using the Adam optimizer \cite{adam}. See the Appendix for more details.



\section{Experiments}

\begin{figure*}[ht]
	\begin{center}
		\includegraphics[width=\linewidth]{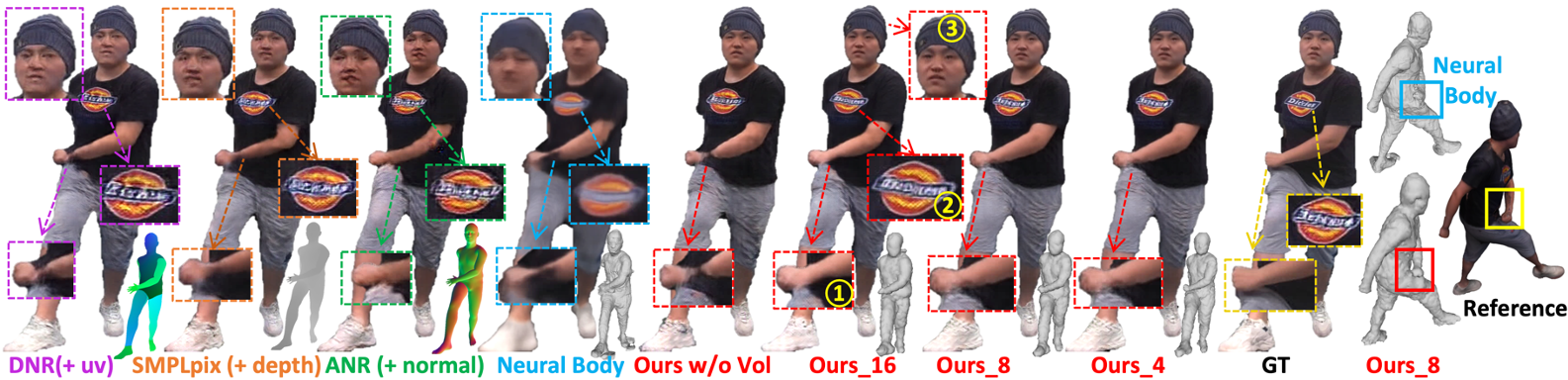}
	\end{center}
	\vspace{-0.2in}
	\caption{Qualitative results of our variants by changing downsampling factor \textbf{\textit{S}} of PD-NeRF. Though existing SMPL based 2D-Plus GAN methods take as input extra geometry priors, such as DNR(+ UV)\cite{dnr}, SMPLpix(+ depth)\cite{smplpix}, ANR(UV + normal)\cite{anr}, they fail to fully utilize the priors for geometry-aware rendering. Instead, ours can handle self-occlusions better \cnum{1} and also improve the rendering quality (\cnum{2}\cnum{3}) by learning a $1/16$ downsampled PD-NeRF (Ours\_16, $45\times45$). For ours and Neural Body, the learned geometries are shown on the right. Note the gap (red) which indicates more accurate geometry reconstruction and encourages the occlusion solving (i.e., the left arm in front of the body).}
	\label{fig:ab_all}
	\vspace{-0.16in}	
\end{figure*}

\noindent \textbf{Dataset}. We evaluate our method on 10 sequences, denoted R1-6, Z1-3, and M1. We captured R1-6, and each dataset has 5 cameras at a resolution of $1280 \times 720$ with 800-2800 frames. Z1-3 from ZJU\_MoCap~\cite{neuralbody} have 24 cameras (1024$\times$1024, 620-1400 frames each), and we use splits of 10/7, 12/8, 5/5 separately for training/test cameras. M1~\cite{Habermann2021RealtimeDD} has 101 cameras (1285$\times$940, 20K frames each), and we utilize 19/8 training/test cameras. For these sequences, we select key sequences to include various motions and use a split of 80\%/20\% for training and testing. All the tested poses are novel, and test viewpoints for R4, Z1-Z3, M1 are new. 
See R1 in Fig.~\ref{fig:ab_all}, R2-R4, Z1, Z3 in Fig.~\ref{fig:rgb5_zju}, M1 in Fig.~\ref{fig:dress}, and the others in the Appendix.


\label{sec:baseline}
\noindent \textbf{Baselines}. We compare our method with GAN-based methods (DNR\cite{dnr}, SMPLpix\cite{smplpix}, ANR\cite{anr}), and Volume Rendering method Neural Body~\cite{neuralbody}. For fair comparisons, DNR, ANR, SMPLpix all have the same generators as ours, the same SMPL model as input, and were trained with the losses mentioned in their papers. {ANR}: Since the code of ANR was not released when this work was developed, we cannot guarantee our reproduced ANR achieves the performance as expected, though it converges and generates reasonable results. 
{SMPLpix}: We follow the author's recent update\footnote{https://github.com/sergeyprokudin/smplpix} to strengthen SMPLpix by rasterizing the SMPL mesh instead of the SMPL vertices~\cite{smplpix}. We optimize a texture latent in the reproduction of SMPLpix to make it drivable by SMPL meshes, and in this paper, we employ the same texture latent $Z_T$ as ours in the reproduction of DNR, ANR and SMPLpix, and all these methods are driven by SMPL meshes. {Neural Body}\cite{neuralbody} was trained with the provided code and setup.   
 

Though Neural Actor (NA \cite{neuralactor}) also conditions a NeRF on the texture features in UV space, whereas ours is distinguished by (1) encoding both geometric and textural features in Pose Encoding, (2) efficient hybrid rendering.

\noindent \textbf{Notation.} Ours\_\textbf{\textit{S}}(\textbf{\textit{N}}) is the variant of our method, where \textbf{\textit{S}} is the downsampling factor of PD-NeRF, and \textbf{\textit{N}} is the number of sample points along each ray. By default, we use the setting of Ours\_8(12) for comparisons in the paper.

\begin{table}[t]
	\begin{center}
	\begin{tabular}{lcccc}
		\hline
		Mean & LPIPS $\downarrow$ & FID $\downarrow$ & SSIM$\uparrow$ & PSNR$\uparrow$\\ \hline
		DNR & .113          & 85.75          & .823          & 25.03 \\         
		SMPLpix & .110          & 80.71          & .826          & 25.15  \\         
		ANR & .127          & 92.38          & .821          & 25.30  \\ 
		NB & .210         & 149.92          & .830          & 25.61 \\
		Ours & \textbf{.100 } & \textbf{72.14} & \textbf{.836} & \textbf{25.68} \\ \hline 
	\end{tabular}	
	\vspace{-0.1in}
	\caption{Quantitative comparisons (averaged on all the nine sequences (R1-6, Z1-3) ). SMpix: SMPLpix, NB: Neural Body. To reduce the influence of the background, all scores are calculated from images cropped to 2D bounding boxes. LPIPS \cite{lpip} and FID \cite{Heusel2017GANsTB} capture human judgement better than per-pixel metrics such as SSIM \cite{ssim} or PSNR. The results for each sequence can be found in the Appendix.}
	\vspace{-0.26in}
	\label{tab:quan_all}
	\end{center}
\end{table}

\subsection{Evaluations}
\noindent \textbf{Differences to the Baseline Methods}. As shown in Fig.~\ref{fig:ab_all}, compared with 2D-Plus methods (DNR, SMPLpix, ANR), we can handle self-occlusions better and generate more details than Neural Body. We also compare the architecture of ours, DNR, and Neural Body in Fig.~\ref{fig:cmp_vr_nr}.

\noindent \textbf{Comparisons}. We evaluate our methods on the 10 sequences, shown in Fig.~\ref{fig:ab_all}, \ref{fig:dress},  \ref{fig:rgb5_zju} (see R5, R6, Z2 in the Appendix). We summarize the quantitative results in Tab.~\ref{tab:quan_all}, where we achieve the best performances on 34/40 evaluations metrics, and on all the 20 LPIPS/FID scores.

\noindent \textbf{Rendering Loose Clothing}. Our method is capable of rendering loose clothing like skirts with GAN as shown in Fig.~\ref{fig:dress}. We visualize the reconstructed geometries and referenced SMPL meshes in Fig. \ref{fig:skirt_geo}. In the current setup (learning PD-NeRF from 64$\times$96 images, e.g., regressed PD-NeRF image \cnum{1}), we cannot reconstruct the full offsets of the skirt, though we recover some offset details \cnum{4} and hair knot \cnum{2}\cnum{3}. Yet we can use GAN to render the skirt with rough geometries as used in \cite{stylepeople}, and geometry learning also enables better temporal consistency than DNR (see demo video). 
However, most animatable NeRF methods \cite{neuralactor,Peng2021AnimatableNR,narf,Chen2021AnimatableNR} parameterize clothes with skinning in a backward warping step, can only handle clothing types that roughly follow the topological structure of the SMPL, but cannot handle skirts.

\begin{table}[t]
	\begin{minipage}{\linewidth}
		\begin{center}
			\includegraphics[width=.95\linewidth]{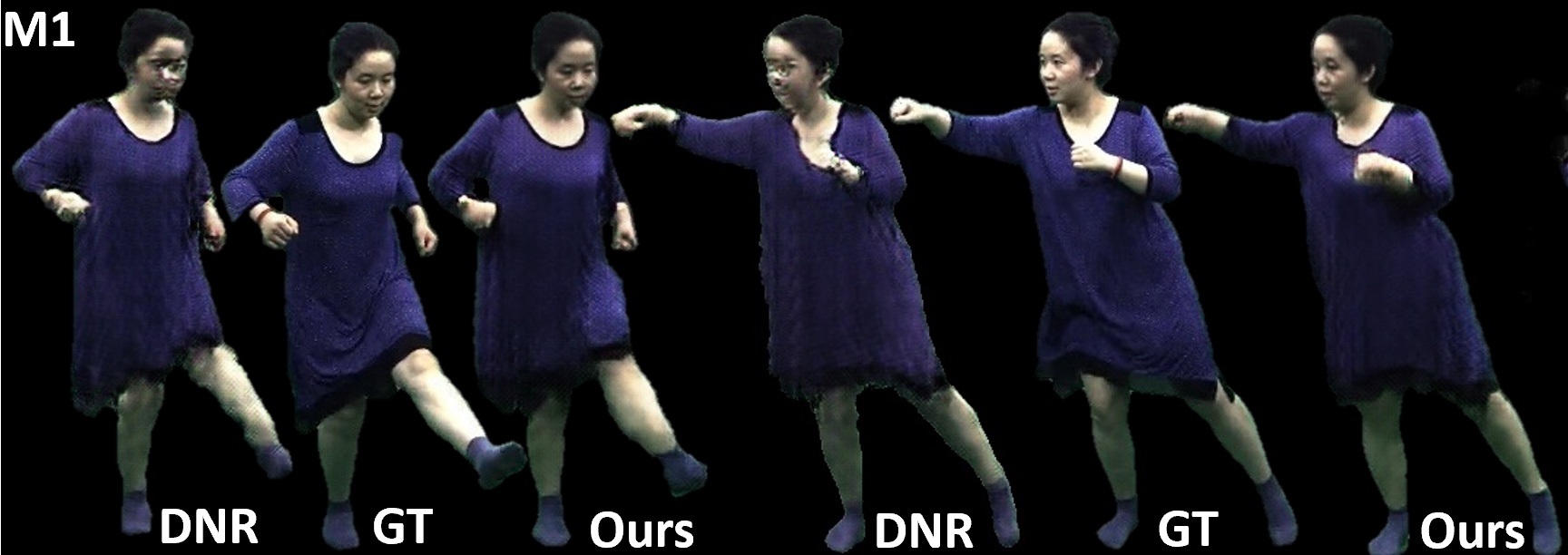}
		\end{center}
		\vspace{-0.18in}
		\captionof{figure}{Render skirts on novel poses and viewpoints.}
		\label{fig:dress}
	\end{minipage}

	\vspace{0.08in}
	\hfill

	\begin{minipage}{\linewidth}
		\begin{center}
			\includegraphics[width=.95\linewidth]{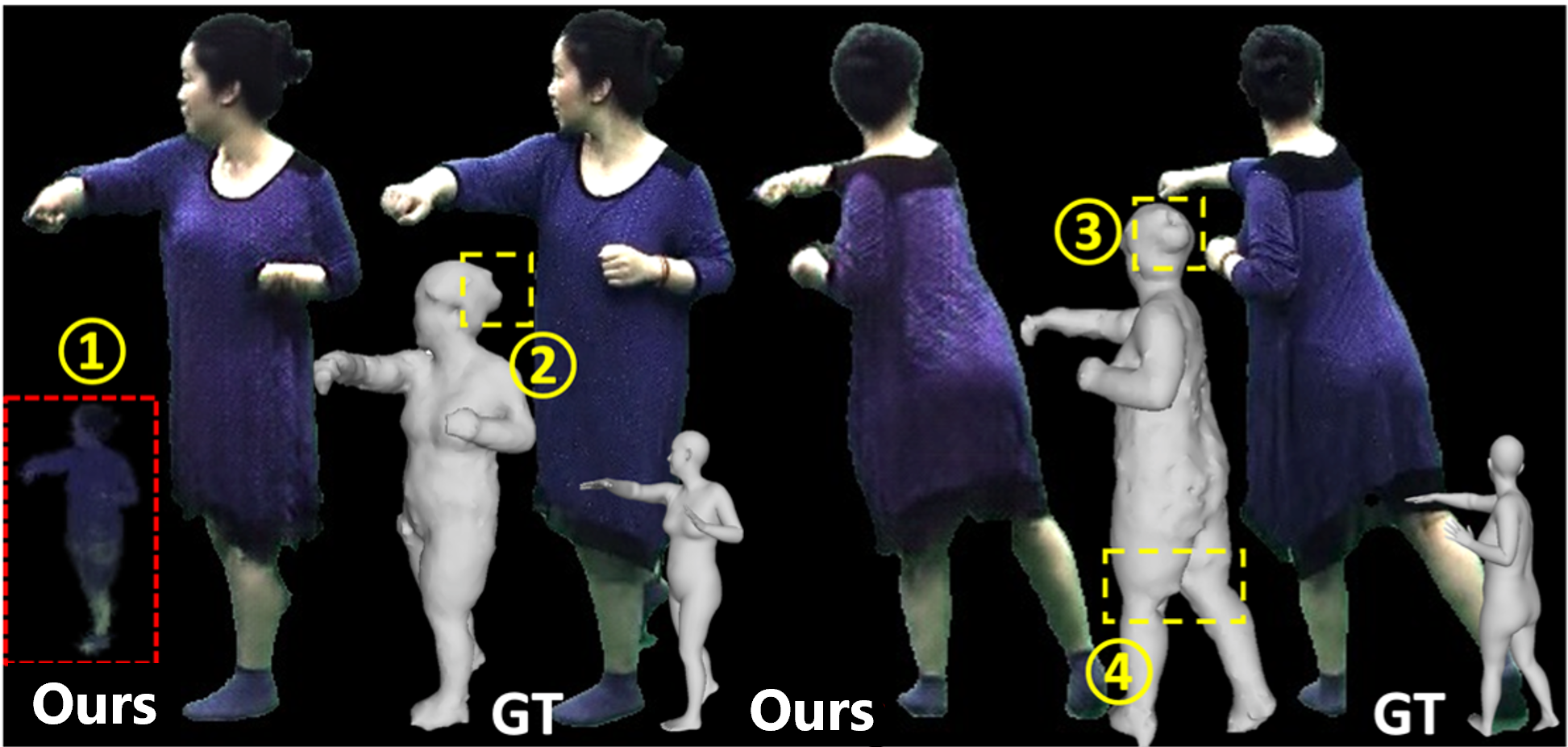}
		\end{center}
	   \vspace{-0.18in}
		\captionof{figure}{Geometry reconstructions of skirt (M1 sequence).}
		\label{fig:skirt_geo}
	\end{minipage}

	\vspace{0.08in}
	\hfill

	\begin{minipage}{\linewidth}
		\begin{center}
			\begin{tabular}{lcccc}
				\hline
				M1 & LPIPS $\downarrow$ & FID $\downarrow$ & SSIM$\uparrow$ & PSNR$\uparrow$\\ \hline
				DNR  & .195          & 144.78          & .687          & 19.96          \\		
				Ours & \textbf{.179} & \textbf{132.83} & \textbf{.696} & \textbf{20.18} \\ \hline
			\end{tabular}
			\vspace{-0.06in}
			\captionof{table}{Quantitative comparisons on M1 sequence.}
			\label{tab:skirt_quant} 
			\vspace{-0.16in}
		\end{center}    
	\end{minipage}

\end{table}

\begin{table}[t]
	\begin{minipage}{\linewidth}
		\begin{center}
			\includegraphics[width=\linewidth]{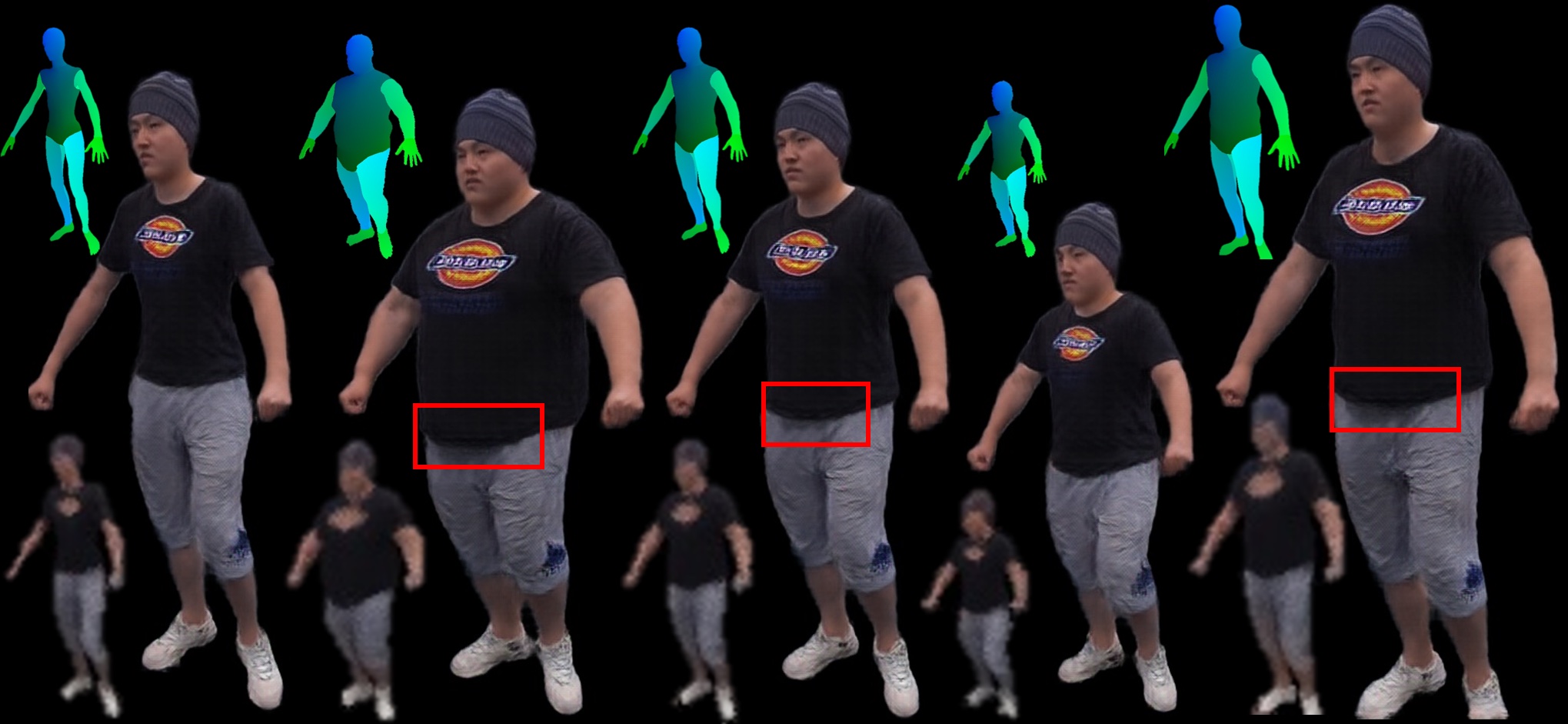}
		\end{center}
		\vspace{-0.18in}
		\captionof{figure}{Rendering results of HVTR for different body shapes of the same individual. Top-left: SMPL shapes (visualized as UV coordinate maps); Bottom-left: renderings of PD-NeRF.}
		\label{fig:shape_edit}
	\end{minipage}
	\vspace{0.1in}
	\hfill

	\begin{minipage}{\linewidth}
		\begin{center}
			\begin{tabular}{lcccc}
				\hline
				Models       & LPIPS & FID     & Time (s) & \multicolumn{1}{c}{VR\_T(\%)} \\ \hline
				DNR          & .102 & 75.02  & .184     & -                             \\
				SMPLpix      & .100 & 69.81  & .198     & -                             \\
				ANR          & .117 & 78.50  & .224     & -                             \\
				Neural Body  & .212 & 155.84 & 18.20   & -                             \\				
				Ours\_16(7)  & .097 & 64.87  & .292     & 11.99                         \\
				Ours\_16(12) & .091 & 64.09  & .295     & 12.88                         \\
				Ours\_16(20) & .091 & 62.88  & .305     & 15.41                         \\
				Ours\_8(12)  & .090 & 62.33  & .349     & 26.36                         \\
				Ours\_4(20)  & \textbf{.086} & \textbf{60.79}  & .464     & 44.61  \\ \hline			
			\end{tabular}
			\vspace{-0.06in}
			\caption{Accuracy and inference time. VR\_T(\%) indicates the percentages of the volume rendering time. We test the end-to-end inference time on a GeForce RTX 3090, and the time for rendering the required maps  are also counted, such as DNR (UV coord maps), SMPLpix (depth maps), ANR (UV coord + normal maps), ours: UV coord + depth maps (used in PD-NeRF to sample query points). PyTorch3D \cite{ravi2020pytorch3d} is used for rendering. Experiments conducted on R1 sequence under novel poses.}
			\label{tab:full_acc_time}
		\end{center}
	\end{minipage}

	\vspace{0.08in}
	\hfill
	
	\begin{minipage}{\linewidth}
		\begin{center}
			\begin{tabular}{lcccc}
				\hline
				& LPIPS $\downarrow$ & FID $\downarrow$ & SSIM$\uparrow$ & PSNR$\uparrow$\\ \hline
				Ours w/o Tex & .106	& 76.43	& .827	& 25.60 \\
				Ours w/o Vol  & .099  & 70.53   & .837  & 25.98 \\			
				NeRF + Pose		& .099	& 68.25	&\textbf{.843}	&\textbf{26.27} \\ \hline
				Ours  & \textbf{.090} & \textbf{62.33} & .842 & 26.22 \\ \hline
			\end{tabular}
			\vspace{-0.06in}
			\caption{Ablation of each component. Ours refers to Ours\_8(12), and all the variants are evaluated under the setting of \textbf{\textit{S=8}} and \textbf{\textit{N=12}}. Experiments conducted on R1 sequence under novel poses.}
			\vspace{-0.2in}
			\label{tab:ab_comp}
			\end{center}
	\end{minipage}
\end{table}

\noindent \textbf{Accuracy, Inference Time}. See the accuracy and inference time in Tab.~\ref{tab:full_acc_time}. We improve the performance over DNR and SMPLpix by about 10\% (even 14\% by Ours\_4) at a small computational cost, and is almost $52\times$ faster than Neural Body. For fair comparisons, we evaluate Tab.~\ref{tab:full_acc_time} on R1 (Fig.~\ref{fig:cmp_vr_nr}) dataset (about 8K frames for training, 2K for testing), where each frame was cropped to $720\times720$ close to the human bounding box (bbox).

\noindent \textbf{Applications}. We can render avatars under user-controlled novel views, poses, and shapes for \textbf{Novel View Synthesis}, \textbf{Animation}, and \textbf{Shape Editing} (Fig.~\ref{fig:shape_edit}). In Fig.~\ref{fig:shape_edit}, we see both PD-NeRF and HVTR generate reasonable results. Not just a straightforward texture to shape mapping, HVTR can generate some shape-dependent wrinkles (marked in red for big models), though these shapes were not seen in training.

\begin{table*}[t]
	\centering
	\begin{minipage}{.95\linewidth}	
	\vspace{-0.12in}
	\begin{center}
		\includegraphics[width=\linewidth]{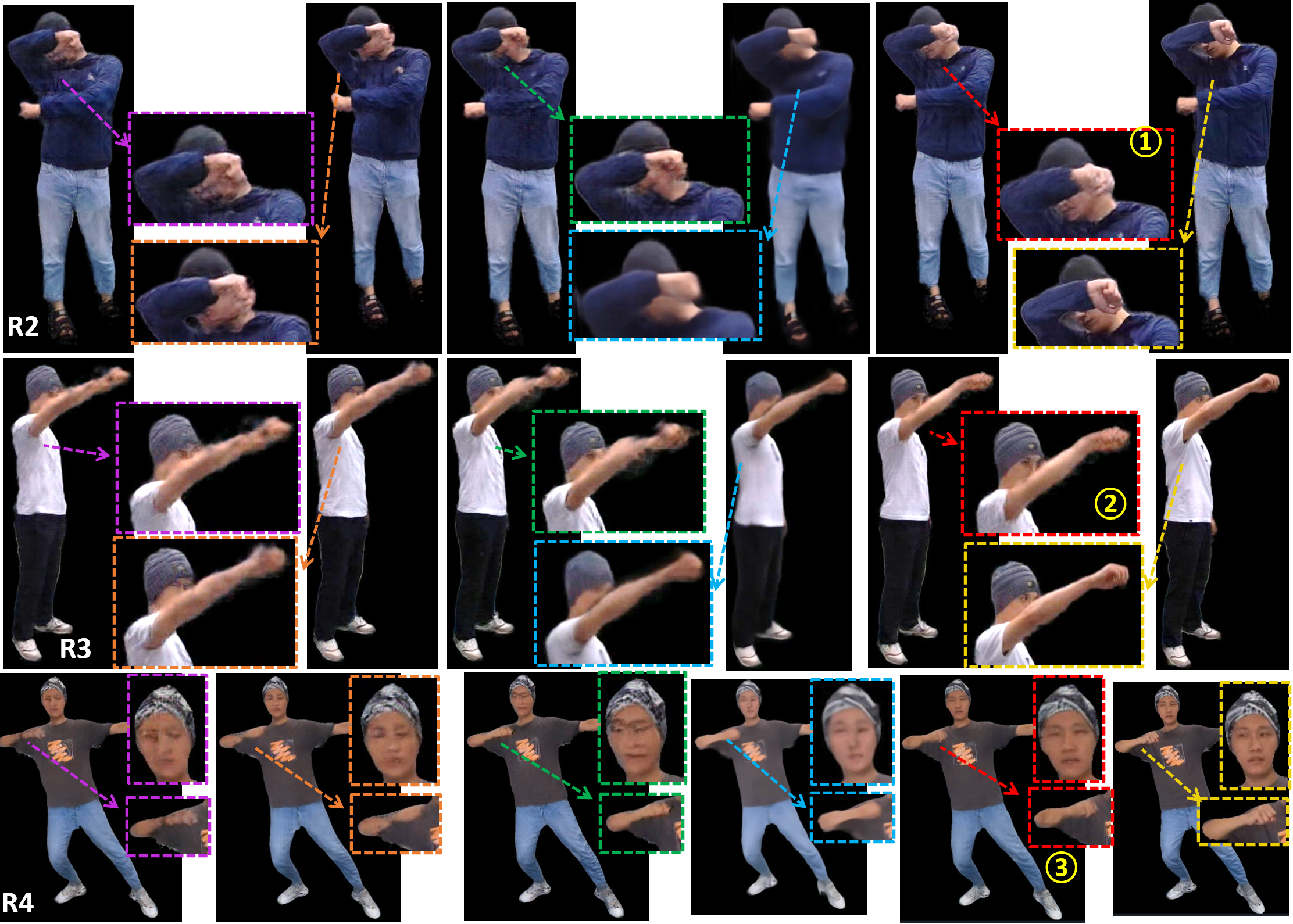}
	\end{center}
	\vspace{-0.13in}
	\end{minipage}
	\hfill
	\begin{minipage}{.95\linewidth}	
		\begin{center}
			\includegraphics[width=\linewidth]{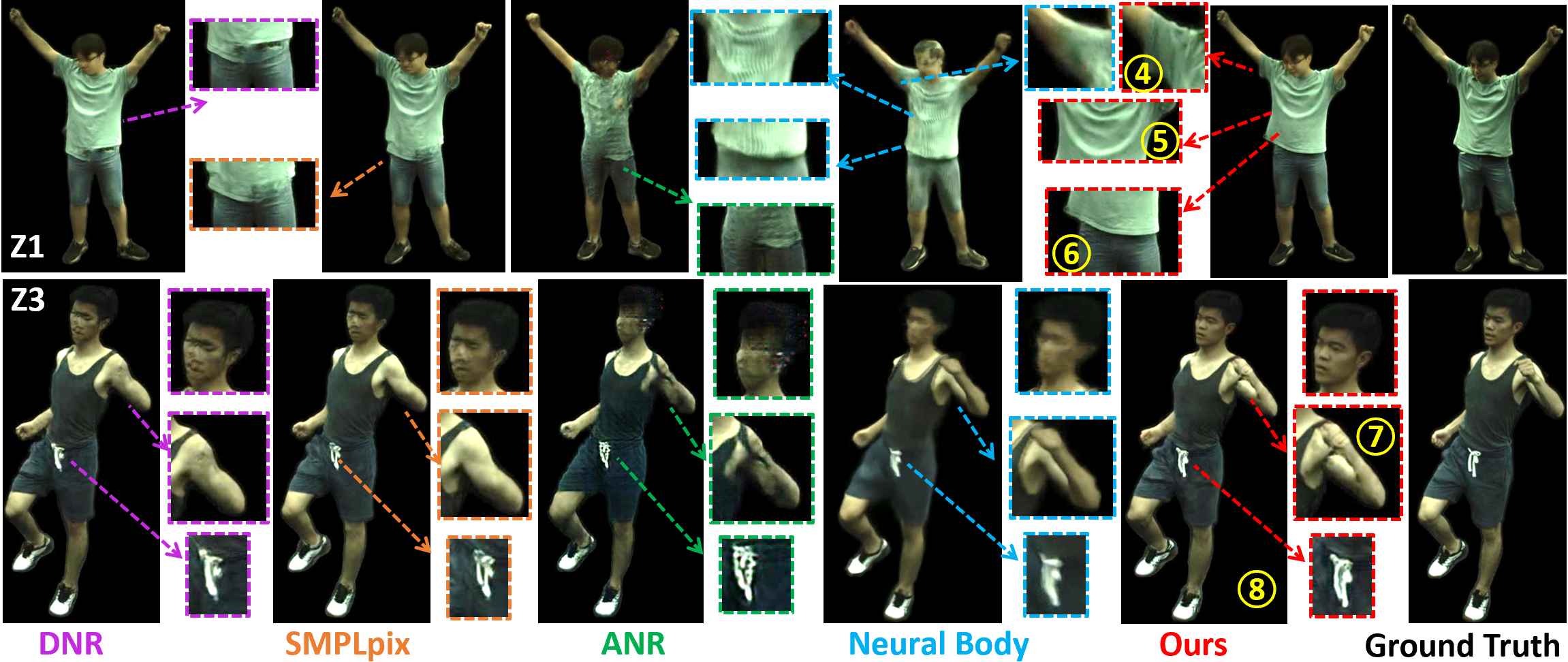}
		\end{center}
		\vspace{-0.12in}
		\captionof{figure}{Comparisons with GAN-based methods (DNR\cite{dnr}, SMPLpix \cite{smplpix}, ANR \cite{anr}), and Neural Body \cite{neuralbody} on R2-4, Z1, and Z3. Our method can generate different levels of pose-dependent details: \cnum{6} offsets, \cnum{5} big wrinkles, \cnum{4} tiny wrinkles. We handle self-occlusions better (\cnum{1}\cnum{2}\cnum{3}\cnum{7}) compared to GAN-based methods, generates high-quality details(\cnum{4}\cnum{5}\cnum{8}), and preserves thin parts (\cnum{3}\cnum{7}) and facial details better. All the poses are novel, and R4, Z1, Z3 are novel views. Note that we cannot guarantee our reproduced ANR achieves the expected performance as stated in Sec.~\ref{sec:baseline}.}
		\vspace{-0.2in}
		\label{fig:rgb5_zju}
	\end{minipage}
\end{table*}

\subsection{Ablation Study}


\noindent \textbf{The effectiveness of Volumetric Learning (PD-NeRF).}  We analyze how PD-NeRF influences the final rendering quality and inference time by evaluating two parameters: the resolution represented by a downsampling factor \textit{\textbf{S}}, and the number of sampled points \textit{\textbf{N}} along each ray, as shown in Fig.~\ref{fig:ab_all}, Tab.~\ref{tab:full_acc_time}. Fig.~\ref{fig:ab_all} shows that we can improve the capability of solving self-occlusions by just incorporating a {$1/16$} ($45\times45$) downsampled PD-NeRF (Ours\_16 vs Ours w/o Vol). In addition, the quantitative results can be further improved by decreasing \textit{\textbf{S}} (e.g., Ours\_8(12) vs. Ours\_16(12)) or sampling more points (e.g., Ours\_16(12) vs. Ours\_16(7)), which demonstrates how PD-NeRF influences the rendering quality. The performances of each variant of our method in Tab. \ref{tab:full_acc_time} vary with the resolution of the human bbox and the ratio of human bbox to the full rendered image, and the variant Ours\_16(20) may fail if the resolution of the human bbox is too low. In constrast, Ours\_8(12) shows better generalization capability on different sequences and this is why we use the setting of Ours\_8(12) in the paper.
 


\noindent \textbf{Ours w/o Tex or Vol.} However, the performances degraded largely when the Texture Encoding part (Ours w/o Tex \cnum{2} in Fig.~\ref{fig:overview}) or Volumetric Representation (Ours w/o Vol \cnum{3}) was removed  (see Tab. \ref{tab:ab_comp}).

\noindent \textbf{UV-based pose parameterization.} We also evaluate a variant (NeRF + Pose) where the NeRF is conditioned on global pose parameters, as listed in Tab. \ref{tab:ab_comp}. Our UV-based parameterization encodes semantics and enables better pose generalization than the global pose parameterization.

\section{Discussion and Conclusion}

\noindent \textbf{Conclusion}. We introduce Hybrid Volumetric-Textural Rendering (HVTR), a novel neural rendering pipeline, to generate human avatars under user-controlled poses, shapes and viewpoints. HVTR can handle complicated motions, render loose clothing, and provide efficient inference. The key is to learn a pose-conditioned downsampled neural radiance field to handle changing geometry, and to incorporate both neural image translation and volume rendering techniques for efficient geometry-aware rendering. We see our framework as a promising component for telepresence. 

Yet one limitation is that the rendered images suffer from appearance flickering or chessboard effects, which  occur in GAN-based renderer (e.g., \cite{dnr,smplpix,anr}) and become more obvious for dynamic humans due to the uncertainties of wrinkles in a long range of motions. Yet, our experimental results show that the image quality can be improved by adding more geometric features (i.e., higher resolution PD-NeRF), as shown in Tab. \ref{tab:full_acc_time}, Fig. \ref{fig:ab_all}.

  
\appendix
\renewcommand{\thesection}{\Alph{section}}

\begin{table*}[th]

	\begin{minipage}{.48\linewidth}		
		\begin{center}
			\includegraphics[width=.9\linewidth]{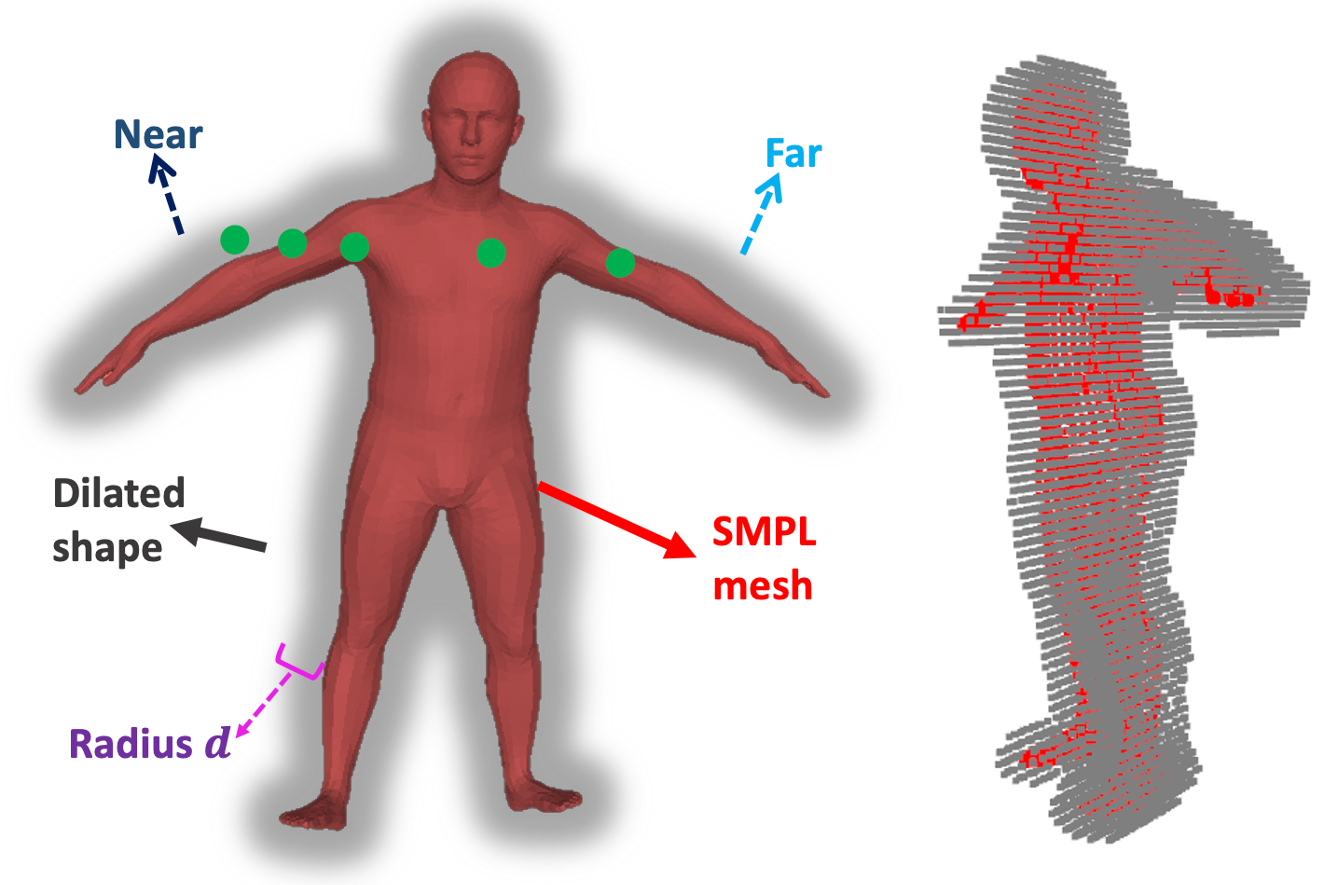}
		\end{center}
		\vspace{-0.12in}
		\captionof{figure}{Geometry-guided ray marching. Left: sampling points by SMPL mesh dilation. Right: Red - SMPL model; Gray - rays and sampled points.}
		\label{fig:dilation}
	\end{minipage}
	\hfill
	\begin{minipage}{.48\linewidth}
		\begin{center}
			\includegraphics[width=\linewidth]{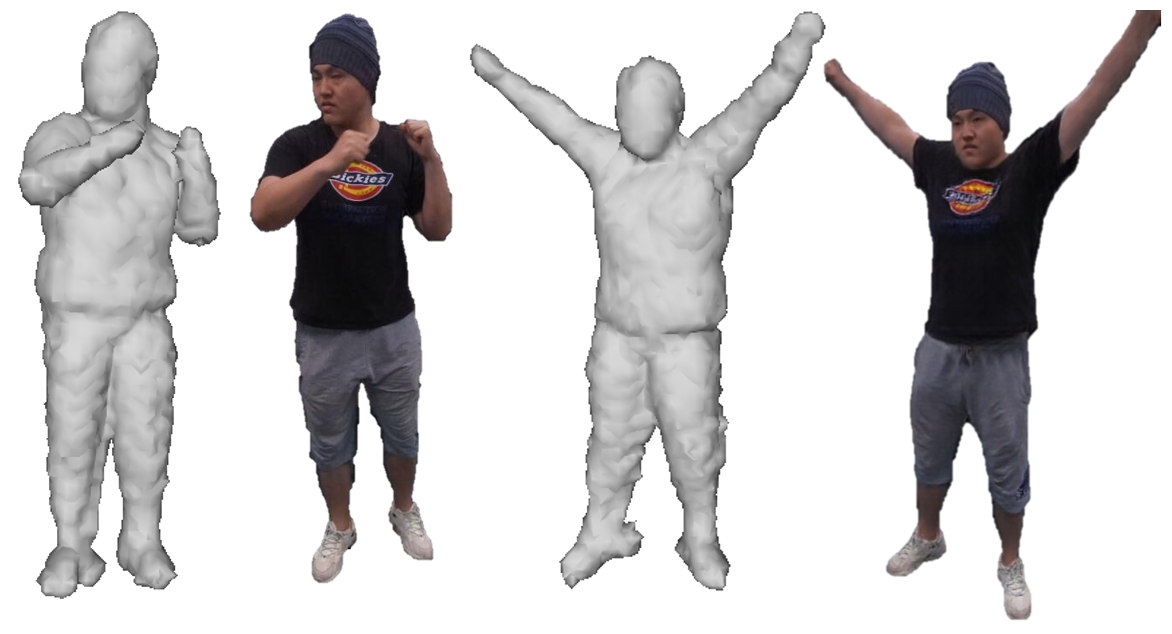}
		\end{center}
		\captionof{figure}{Construct pose-conditioned NeRF with $45\times45$ resolution images and 7 sampled point along each ray: left (geometry), right (reference image).}
		\label{fig:geo45}
	\end{minipage}

	\vspace{0.2in}

	\hfill
	
	\begin{minipage}{\linewidth}
		\begin{center}
			\includegraphics[width=.95\linewidth]{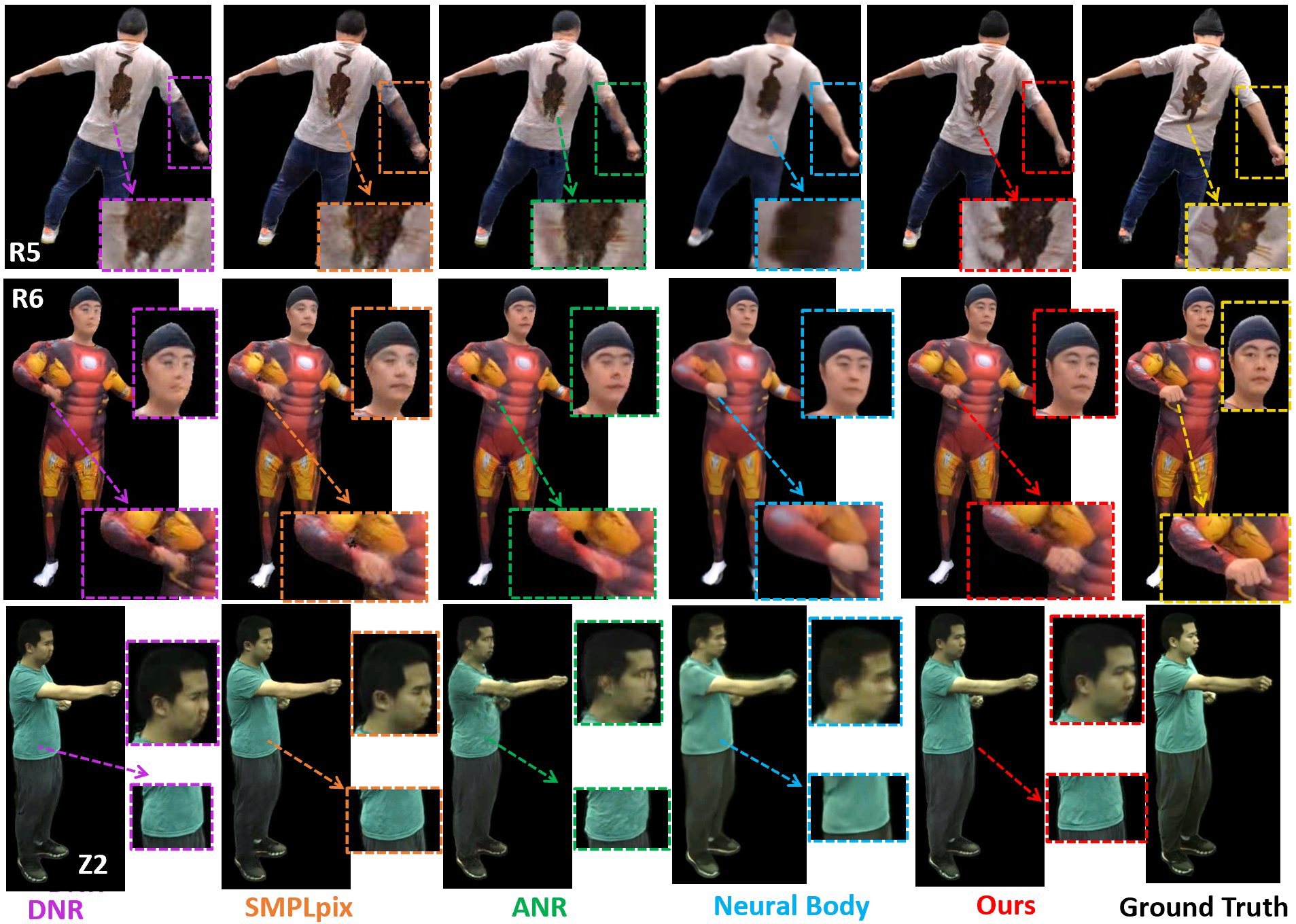}
		\end{center}
		\captionof{figure}{Qualitative results on unseen sequences on R5, R6, and Z2.}
		\label{fig:otherhuman}
	\end{minipage}
\end{table*}


\begin{table*}[th]

	\centering
	\begin{minipage}{.7\linewidth}
		\begin{center}
			\includegraphics[width=\linewidth]{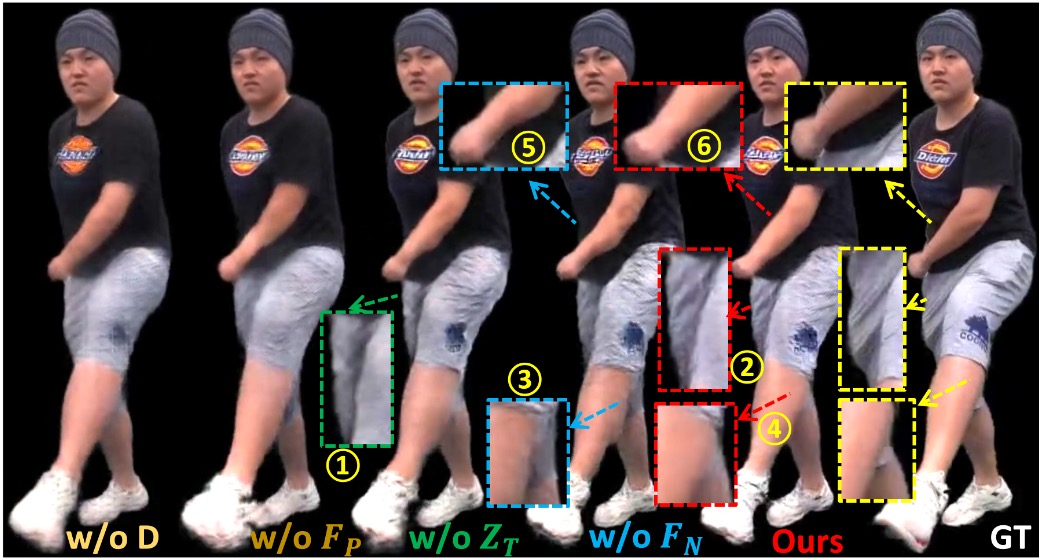}
		\end{center}    
	\end{minipage}
	
	\hfill
	\vspace{0.2in}

	\begin{minipage}{.8\linewidth}
		\begin{center}
			\begin{tabular}{lcccc}
				\hline
				& LPIPS $\downarrow$ & FID $\downarrow$ & SSIM$\uparrow$ & PSNR$\uparrow$\\ \hline
				w/o D       & 1.210  & 87.533 & \textbf{.850} & \textbf{26.863} \\
				w/o $F_P$   & .955  & 67.613 & .838 & 26.167 \\
				w/o $F_N$   & .927  & 65.310 & .839 & 26.295 \\
				w/o $F_T$   & .970  & 69.063 & .835 & 26.027 \\
				w/o $Z_G$   & .904  & 63.114 & .843 & 26.310 \\        
				w/o $Z_T$ & .942	& 63.413	&\textbf{.850}	& {26.551} \\ 
    
				Ours        & \textbf{.901}  & \textbf{62.333} & .842 & 26.217 \\ \hline
			\end{tabular}


			\captionof{figure}{Ablation study of each component. `w/o D': variant without discriminator. Discriminator and adversarial training enforce realistic rendering (e.g., ours vs. `w/o D'). PoseNet $F_P$ maps the input poses to a higher dimensional  space, which enables better fitting of data that contains high frequency variation, similar to the positional encoding in NeRF \cite{mildenhall2020nerf}, whereas `w/o $F_P$' leads to blurry results. Normal supervision enables our methods to handle shapes (\cnum{6} vs. \cnum{5}) and blend boundaries (\cnum{4} vs. \cnum{3}) better (see ours vs. `w/o $F_N$'). Texture latent $Z_T$ helps capture local details (e.g., \cnum{2} vs. \cnum{1}). Ablation study conducted under the same setting as Tab. \ref{tab:ab_comp}.}
			\label{fig:ab_comp_all}
		\end{center}    
	
	\end{minipage}

	\hfill
	\vspace{0.14in}

	\begin{minipage}{.8\linewidth}
		\begin{center}
			\includegraphics[width= \linewidth]{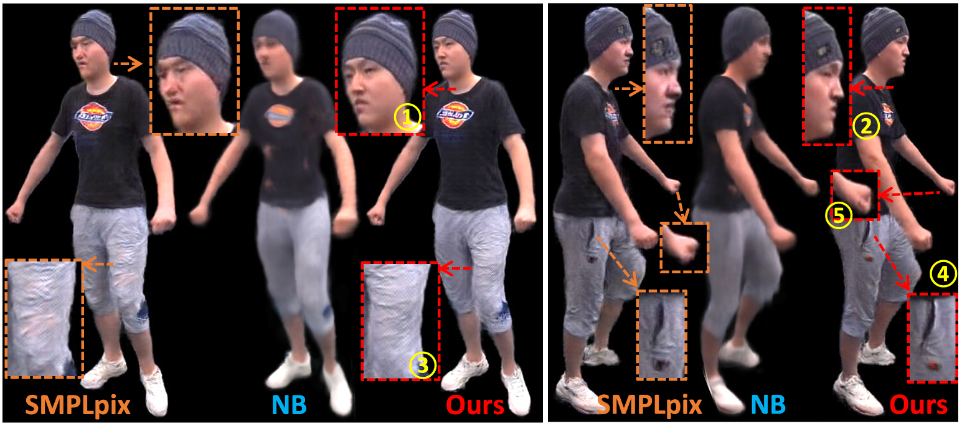}
		\end{center}

		\captionof{figure}{Comparisons on shape editing. Our methods generate high-quality faces \cnum{1}\cnum{2}, wrinkles \cnum{3}\cnum{4} and shapes \cnum{5}, and disentangle body and clothes better \cnum{3} than SMPLpix.}
		\label{fig:shapeedit_comp}
		\vspace{-0.14in}
	\end{minipage}

\end{table*}

\clearpage
\newpage

\clearpage
\newpage

\newcommand*{\appendixdef}{\scalebox{1.2}{\textbf{Appendix.}}}
\noindent \appendixdef

\begingroup
\renewcommand{\arraystretch}{1.25}
\begin{table}[h]
	\centering

	\begin{tabular}{cl}
		\hline
		Term                       & Description                                      \\ \hline \hline
		$M_P$                       & Input SMPL mesh                                  \\
		$o$, $d$                        & Target viewpoint (camera position and direction) \\
		$I_{t}$               & Rendered target image                            \\ \hline \hline
		$F_N$                       & NormalNet (network)                              \\
		$F_P$                       & PoseNet (network)                                \\
		$F_R$                       & TexRenderer (network)                            \\
		$F_T$                       & Texture Encoder (network)                        \\
		$F_\theta$                       & MLPs of PD-NeRF (network)                        \\
		$Z_G$                       & Optimizable Geometry Latent                      \\
		$Z_T$                       & Optimizable Texture Latent                       \\ \hline \hline
		$\Psi^{uv}_{N}$  & Normal of SMPL mesh in UV space           \\
		$\Psi^{uv}_{P}$  & Stacked pose-dependent featuers in UV space      \\
		$\Psi^{3D}_{P}$ & Features of PD-NeRF in 3D space                  \\
		$\Psi^{im}_{P}$ & Pose-dependent neural textures in image space    \\
		$\Psi^{im}_{tex}$ & Extracted textural features in image space       \\
		$\Psi^{im}_{vol}$ & Rasterized volumetric features in image space    \\
		$\Psi^{im}_{uv}$ & UV coordinate map in image space   \\ \hline            
	\end{tabular}
	\caption{Glossary table. Terms that we used in the paper.}
	\label{tab:term_table}
\end{table}
\endgroup

\section{More Experimental Results}
\subsection{Comparisons}
    
\noindent \textbf{Free-Viewpoint Synthesis.} We evaluate the results of image synthesis on novel views while the test poses are the same as training poses and follow the same protocol as Neural Body \cite{neuralbody} on Z1 and Z3 sequences from ZJU\_MoCap \cite{neuralbody}. Our method recovers more details, and significantly outperforms Neural Body on LPIPS \cite{lpip} and FID \cite{Heusel2017GANsTB} as shown in Fig. \ref{fig:free_view_nb}.

\begin{table*}[th]\small
	\begin{minipage}{\linewidth}		
		\begin{center}
			\includegraphics[width=\linewidth]{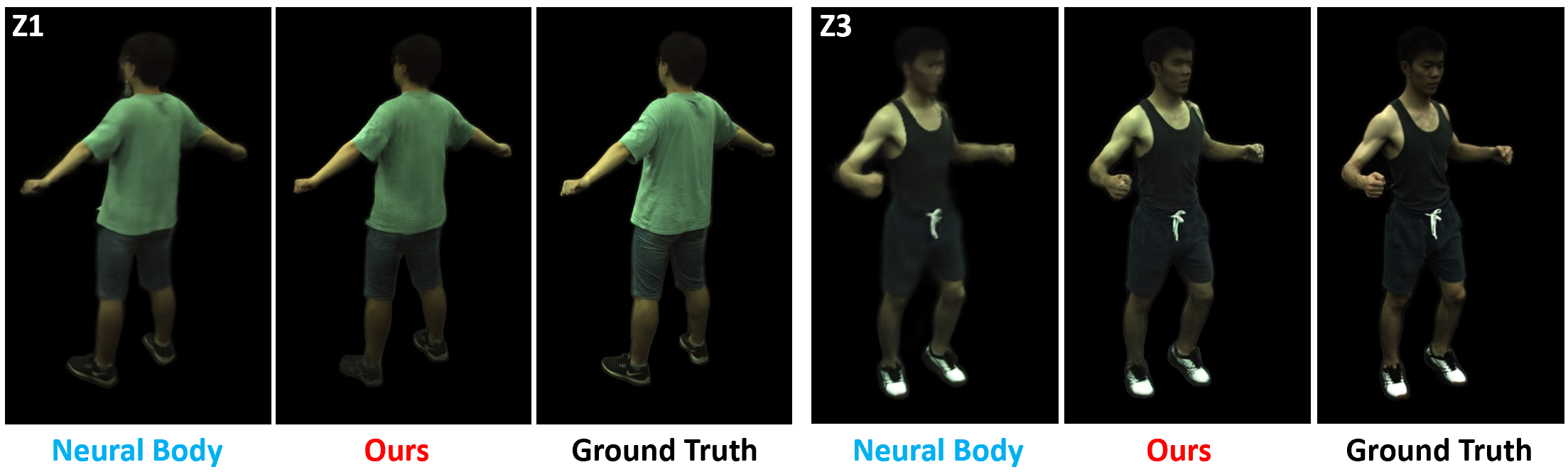}
		\end{center}
	\end{minipage}

	\vspace{0.2in}

	\begin{minipage}{.48\linewidth}
		\begin{center}
			\begin{tabular}{ccccc}
				\hline
				Z1 (this example)  & LPIPS $\downarrow$ & FID $\downarrow$ & SSIM$\uparrow$ & PSNR$\uparrow$\\ \hline
				Neural Body \cite{neuralbody}  & .153          & 90.74          & \textbf{.886}          & \textbf{26.51}          \\
				Ours & \textbf{.112} & \textbf{56.94} & {.855} & {23.75} \\ \hline \hline
				\textbf{Z1 (mean)} & LPIPS $\downarrow$ & FID $\downarrow$ & SSIM$\uparrow$ & PSNR$\uparrow$\\ \hline
				Neural Body \cite{neuralbody}  & .137          & 120.73          & \textbf{.889}          & {\textbf{29.09}}          \\
				Ours & \textbf{.117} & \textbf{87.55} & {.853} & {26.16} \\ \hline
			\end{tabular}
		\end{center}
	\end{minipage}
	\hfill
	\begin{minipage}{.48\linewidth}
		\begin{center}
			\begin{tabular}{ccccc}
				\hline
				Z3 (this example)  & LPIPS  & FID  & SSIM & PSNR\\ \hline 
				Neural Body \cite{neuralbody}  & .170          & 112.70          & \textbf{.866}          & \textbf{26.16}          \\
				Ours & \textbf{.124} & \textbf{52.23} & {.816} & {22.09} \\ \hline \hline
				\textbf{Z3 (mean)} & LPIPS  & FID  & SSIM & PSNR\\ \hline
				Neural Body \cite{neuralbody}  & .149          & 132.05          & \textbf{.872}          & \textbf{26.35}          \\
				Ours & \textbf{.098} & \textbf{69.79} & {.848} & {24.08} \\ \hline
			\end{tabular}
		\end{center}
	\end{minipage}
	\captionof{figure}{Novel view synthesis. Both methods are trained on 4 cameras, and evaluated on 18 cameras.
	}
	\label{fig:free_view_nb} 
\end{table*}

\begin{table}
	\begin{minipage}{\linewidth}
		\begin{center}
			\includegraphics[width=\linewidth]{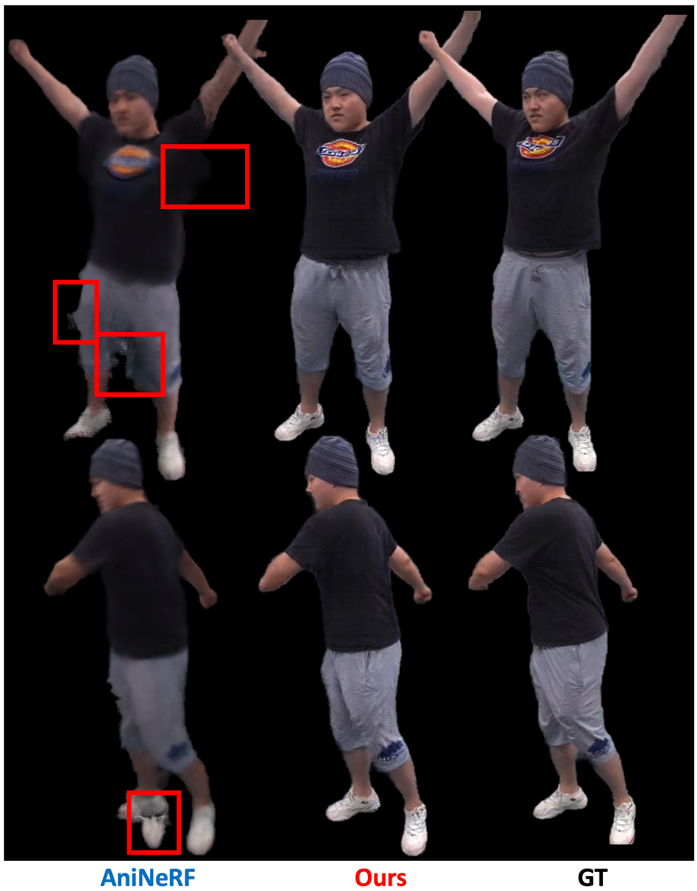}
			\captionof{figure}{Comparisons of backward skinning method (AniNeRF \cite{Peng2021AnimatableNR}) and forward skinning method (ours).  AniNeRF may suffer from artificats due to the one-to-many correspondence problem in backward skinning.}
			\label{fig:cmp_ani}
		\end{center}
	\end{minipage}

	\vspace{0.1in}
	\hfill
	
	\begin{minipage}{\linewidth}
		\begin{center}
			\begin{tabular}{ccccc}
				\hline
				& LPIPS $\downarrow$ & FID $\downarrow$ & SSIM$\uparrow$ & PSNR$\uparrow$\\ \hline
				AniNeRF \cite{Peng2021AnimatableNR}  & .271          & 196.44          & .773          & 23.36          \\
				Ours & \textbf{.090} & \textbf{62.33} & \textbf{.842} & \textbf{26.22} \\ \hline
			\end{tabular}
			\captionof{table}{Comparisons with AniNeRF on R1 sequence under novel poses.}
			\label{tab:aninerf}
		\end{center}
	\end{minipage}
\end{table}

\noindent \textbf{Qualitative Comparisons on R5, R6, and Z2 Dataset.}
The qualitative comparisons on R5, R6, and Z2 are shown in Fig.~\ref{fig:otherhuman}. 

\noindent \textbf{Accuracy and Inference Time.}
The accuracy and inference time of each method are listed in Tab.~\ref{tab:full_acc_time_supp}.

\noindent \textbf{GPU Memory.} Ours\_4(20)-most GPU-consuming version: 21GB; Neural Body: 5GB; ANR:11GB. Note that we need much GPU memory in training, because HVTR renders a full image of PD-NeRF at each iteration, whereas Neural Body just randomly sample points to calculate gradients.  However, in inference, Ours\_4(20): 4GB, Neural Body: 15GB. Note that this was evaluated on the cropped bbox with downsampled \textit{\textbf{S=4}}.

\noindent \textbf{Shape Editing.} Comparisons against SMPLpix \cite{smplpix} and Neural Body \cite{neuralbody} on shape editing are shown in Fig.~\ref{fig:shapeedit_comp}. Our methods generate high-quality faces \cnum{1}\cnum{2} and wrinkles \cnum{3}\cnum{4} and shapes \cnum{5} and disentangle body and clothes better \cnum{3} than SMPLpix \cite{smplpix}.

\noindent \textbf{Forward vs. Backward Skinning}. AniNeRF \cite{Peng2021AnimatableNR} parameterizes clothes with skinning weights by learning backward skinning, which generally requires a large training dataset, and may suffer from one-to-many backward correspondences \cite{Chen2021SNARFDF} as shown in Fig. \ref{fig:cmp_ani}, whereas we only need a forward skinning step and avoid the multi-correspondence problem. The experiment on R1 dataset shows that our method significantly outperforms AniNeRF on all the four metrics in Tab. \ref{tab:aninerf}, where the same protocol as Tab. \ref{tab:quan_all} is used for training and testing.

\subsection{Ablation Study}

\noindent \textbf{The Effectiveness of Each Component}. We evaluate the effectiveness of the discriminator, PoseNet $F_P$, NormalNet $F_N$, Texture Encoder $F_T$, Geometry latent $Z_G$ and Texture latent $Z_T$, and the qualitative and quantitative results can be found in Fig. \ref{fig:ab_comp_all}. In addition, we found normal supervision also work for the skirt case (see Fig. \ref{fig:skirt_normal}), i.e., ours outperforms `w/o $F_N$' with smoother surface reconstructions \cnum{5} vs. \cnum{6} and high-quality face rendering \cnum{7}\cnum{8}.

\noindent \textbf{Face Identity Loss}. Face Identity Loss (also used in \cite{egorend,feanet}) is a perception loss just for face region, and it can improve the visual quality of face, whereas cannot improve the quantitative LPIPS or FID scores of the whole body as listed in Tab.~\ref{tab:face_loss} since: 1) LPIPS/FID cannot totally capture human judgement. 2) The SMPL mesh we used cannot recover the facial expressions, whereas a blurry or averaged face rendering may get lower LPIPS/FID for this case.

\begin{figure*}[ht]
	\begin{center}
		\includegraphics[width=.8\linewidth]{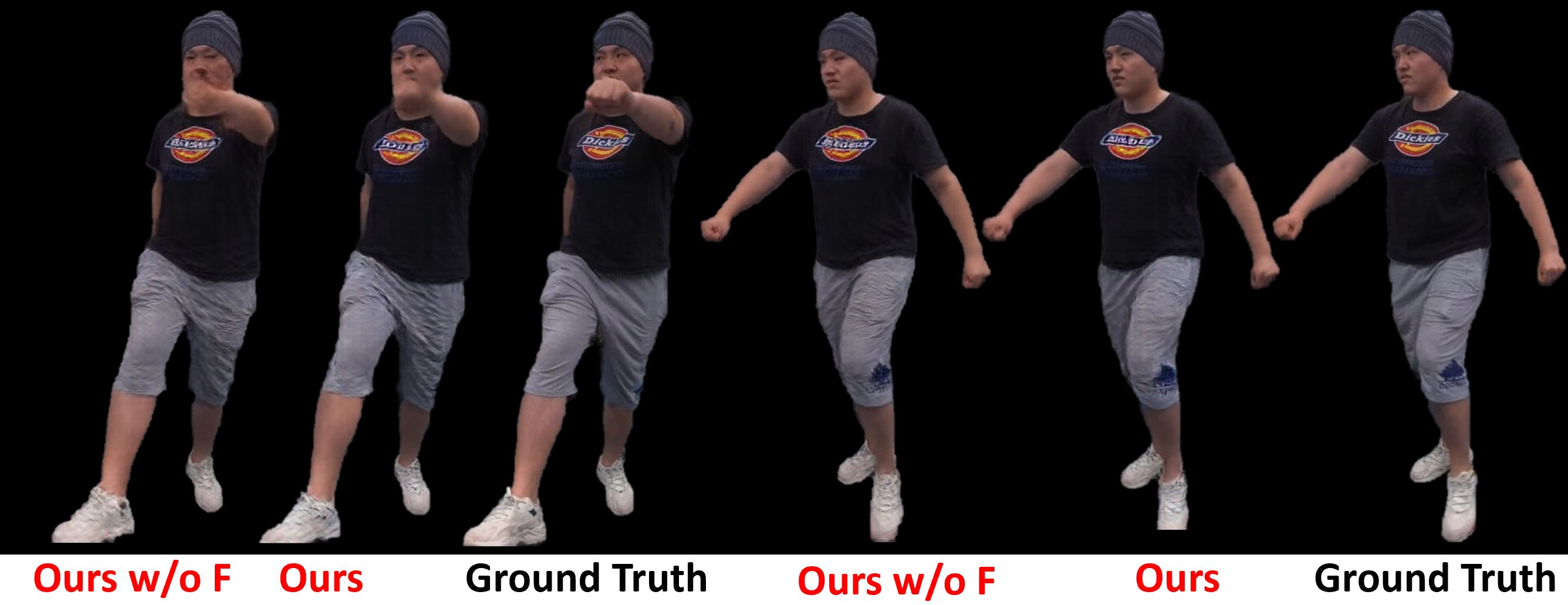}
	\end{center}
	\vspace{-0.16in}
	\caption{Ablation study of face identity loss. Face identity supervision improves the visual quality of face.}
	\label{fig:face}
\end{figure*}

\begin{table}
	\centering
	\begin{tabular}{lcccc}
		\hline
		\textbf{R3} & LPIPS $\downarrow$ & FID $\downarrow$ & SSIM$\uparrow$ & PSNR$\uparrow$\\ \hline
		DNR	w/o F &.108	& 80.33	&.809	&24.05 \\
		DNR w/ F  &\textbf{.103}	&\textbf{75.42}	&\textbf{.812}	&\textbf{24.13} \\ \hline
		
		SMPLPix	w/o F &\textbf{.104}	&\textbf{74.57}	&.810	&\textbf{24.16} \\
		SMPLPix w/ F	&.109	&78.59	&\textbf{.811}	&23.93 \\ \hline \hline
		
		\textbf{R2} & LPIPS $\downarrow$ & FID $\downarrow$ & SSIM$\uparrow$ & PSNR$\uparrow$\\ \hline
		DNR	w/o F & .128	& 105.63	& \textbf{.820}	& 27.82 \\
		DNR w/ F	& \textbf{.125}	& \textbf{98.86}	& \textbf{.820}	& \textbf{27.92} \\ \hline
		
		SMPLPix	w/o F &\textbf{.124}	&99.81	&.822	& 27.92 \\
		SMPLPix w/ F	&\textbf{.124}	&\textbf{94.81}	&\textbf{.826}	& \textbf{27.97} \\ \hline \hline
		
		\textbf{R1} & LPIPS $\downarrow$ & FID $\downarrow$ & SSIM$\uparrow$ & PSNR$\uparrow$\\ \hline
		DNR	w/o F & \textbf{.102}	& \textbf{75.02}	& .831	& 25.73 \\
		DNR w/ F &	.103 &	75.35 &	\textbf{.832} &	\textbf{25.85} \\ \hline
		
		SMPLPix	w/o F & \textbf{.100}	& \textbf{69.81}	& \textbf{.835}	& \textbf{25.93} \\
		SMPLPix w/ F	& .104	& 75.34	& .833	& 25.81 \\ \hline
		
		Ours w/o F	& \textbf{.088}	& \textbf{61.03}	& .841	& \textbf{26.42} \\
		Ours 	&.090	& 62.33	& \textbf{.842}	& 26.22 \\ \hline
	\end{tabular}
	\caption{Quantitative results of each method trained with or without face identity loss. Ours w/o F indicates a variant of our method that is not trained with face identity loss. (DNR w/ F) and (SMPLpix w/ F) are trained with face identity loss.}
	\label{tab:face_loss}
\end{table}	

\noindent \textbf{Feature Fusion}. We compare two methods to fuse the volumetric and textural features as discussed at Sec. \ref{sec:tex_rendering} by concatenation (Concat) and Attentional Feature Fusion (AFF \cite{dai21aff}) on two sequences, R1 and R2 (about 12K frames in training, 3K frames in testing) in Tab.~\ref{tab:feat_fusion}, which shows that AFF can improve the LPIPS \cite{lpip} and FID \cite{Heusel2017GANsTB} results.


\begin{table}

	\begin{minipage}{\linewidth}
		\begin{center}
			\includegraphics[width=.95\linewidth]{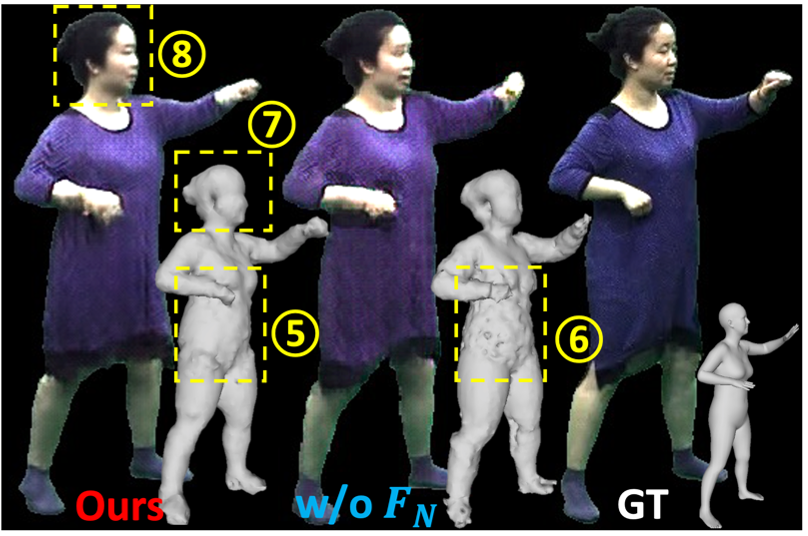}
		\end{center}
	\end{minipage}
	
	\hfill
	\vspace{0.1in}

	\begin{minipage}{\linewidth}
		\begin{center}
			\begin{tabular}{lcccc}
				\hline
				M1 & LPIPS $\downarrow$ & FID $\downarrow$ & SSIM$\uparrow$ & PSNR$\uparrow$\\ \hline
				Ours w/o $F_N$  & .191 & \textbf{130.95} & .685 & 19.97 \\
				Ours & \textbf{.179} & {132.83} & \textbf{.696} & \textbf{20.18} \\ \hline
			\end{tabular}
			\captionof{figure}{Normal prediction also works for skirt with smoother surface reconstructions \cnum{5} vs. \cnum{6} and high-quality face rendering \cnum{7}\cnum{8}.}
			\label{fig:skirt_normal} 
			\vspace{-0.16in}
		\end{center}	
	\end{minipage}

\end{table}
    
\begin{table}
	\begin{center}			
		\begin{tabular}{ccccc}
			\hline
			\textbf{R2} & LPIPS $\downarrow$ & FID $\downarrow$ & SSIM$\uparrow$ & PSNR$\uparrow$\\ \hline
			Concat  & .117          & 90.43          & \textbf{.838}          & {28.55}          \\
			AFF & \textbf{.108} & \textbf{84.43} & {.833} & \textbf{28.62} \\ \hline
			\hline
			\textbf{R1} & LPIPS $\downarrow$ & FID $\downarrow$ & SSIM$\uparrow$ & PSNR$\uparrow$\\ \hline
			Concat  & .099          & 64.54          & \textbf{.856}          & \textbf{26.78}          \\
			AFF & \textbf{.090} & \textbf{62.33} & {.842} & {26.22} \\ \hline
		\end{tabular}
	\end{center}
	\vspace{-0.12in}
		\caption{Comparisons of fusing volumetric and textural features by concatenation (Concat) and Attentional Feature Fusion (AFF \cite{dai21aff}) on R1 and R2 sequence.}
		\label{tab:feat_fusion}
	\end{table} 


\begin{table*}[t]\small
    \centering
	\begin{tabular}{lcccccccccccc}
		\hline
		& \multicolumn{4}{l}{R1}                                          & \multicolumn{4}{l}{R2}                                          & \multicolumn{4}{l}{R3}                                          \\ \hline
		Models      & LPIPS$\downarrow$         & FID$\downarrow$            & SSIM$\uparrow$          & PSNR$\uparrow$           & LPIPS        & FID            & SSIM          & PSNR           & LPIPS         & FID            & SSIM          & PSNR           \\ \hline 
		DNR         & .102          & 75.02          & .831          & 25.73          & .125          & 98.86          & .820          & 27.92          & .108          & 80.33          & .809          & 24.05          \\
		SMpix     & .100          & 69.81          & .835          & 25.93          & .124          & 94.81          & .826          & 27.97          & .104          & 74.57          & .810          & 24.16          \\
		ANR         & .117          & 78.50          & .830          & 26.02          & .129          & 101.72         & .825          & 28.30          & .098          & 69.14          & .813          & 24.29          \\
		NB & .212          & 155.84         & .833          & 26.17          & .218          & 161.99         & \textbf{.833} & 28.61          & .240          & 165.03         & .811          & 24.16          \\
		Ours        & \textbf{.090} & \textbf{62.33} & \textbf{.842} & \textbf{26.22} & \textbf{.108} & \textbf{84.43} & \textbf{.833}          & \textbf{28.62} & \textbf{.093} & \textbf{66.01} & \textbf{.823} & \textbf{24.55} \\
		\hline \hline
		& \multicolumn{4}{l}{R4}                                          & \multicolumn{4}{l}{R5}                                          & \multicolumn{4}{l}{R6}                                          \\
		\hline
		DNR         & .108          & 93.16          & .833          & 23.34          & .136          & 121.50         & .817          & 24.06          & .088          & 74.77          & .864          & 25.81          \\
		SMpix     & .107          & 88.14          & .837          & 23.37          & .131          & 118.64         & .818          & 24.10          & .077          & 64.33          & .875          & 26.14          \\
		ANR         & .138          & 91.92          & .812          & 23.26          & .140          & 123.55         & .823          & 24.67          & .083          & 63.16          & .875          & 26.61          \\
		NB & .198          & 126.26         & \textbf{.856} & \textbf{24.26} & .220          & 161.93         & .816          & 24.25          & .142          & 94.96          & .880          & 27.19          \\
		Ours        & \textbf{.096} & \textbf{78.79} & .849          & 23.98          & \textbf{.117} & \textbf{93.56} & \textbf{.827} & \textbf{24.84} & \textbf{.070} & \textbf{57.00} & \textbf{.891} & \textbf{27.42} \\ \hline \hline
		& \multicolumn{4}{l}{Z1}                                          & \multicolumn{4}{l}{Z2}                                          & \multicolumn{4}{l}{Z3}                                          \\ \hline
		DNR         & .145          & 92.78          & .797          & 22.06          & .145          & 87.27          & .774          & 25.04          & .109          & 82.79          & .826          & 23.16          \\
		SMpix     & .150          & 90.90          & .797          & 22.14          & .144          & 81.78          & .774          & 25.18          & .113          & 83.96          & .827          & 22.92          \\
		ANR         & .205          & 171.69         & .775          & \textbf{22.35}          & .159          & 110.85         & .778          & 25.41          & .173          & 123.84         & .790          & 22.14          \\
		NB & .215          & 163.83         & .789          & 22.16          & .238          & 155.27         & \textbf{.792}          & \textbf{25.88}          & .204          & 167.66         & .825          & \textbf{23.89} \\
		Ours        & \textbf{.143} & \textbf{90.43} & \textbf{.805} & {22.31} & \textbf{.132} & \textbf{79.14} & {.785} & {25.69} & \textbf{.105} & \textbf{78.03} & \textbf{.829} & 23.23          \\ \hline 
	\end{tabular}
	\vspace{0.02in}
	\caption{Quantitative comparisons on nine sequences (averaged on all test views and poses). SMpix: SMPLpix, NB: Neural Body. To reduce the influence of the background, all scores are calculated from images cropped to 2D bounding boxes. LPIPS \cite{lpip} and FID \cite{Heusel2017GANsTB} capture human judgement better than per-pixel metrics such as SSIM \cite{ssim} or PSNR. All poses are novel, and R4, Z1-Z3 are tested on new views.} 
	\label{tab:quan_all}
\end{table*}


\section{Implementation Details}
\noindent \textbf{Optimization}. The networks were trained using the Adam optimizer \cite{adam} with an initial learning rate of $2\times 10^{-4}$, $\beta_1 = 0.5$. The loss weights \{$\lambda_{vol}$, $\lambda_{norm}$, $\lambda_{feat}$, $\lambda_{mask}$, $\lambda_{pix}$, $\lambda_{adv}$, $\lambda_{face}$\} are set empirically to $\{15,1,10,5,1,1,5\}$. We trained DNR \cite{dnr}, SMPLpix \cite{smplpix}, ANR \cite{anr}, and our method for 50,000 iterations, and 180,000 iterations for Neural Body \cite{neuralbody}. The networks were trained on an Nvidia P6000 GPU, and it generally took 28 hours for DNR and SMPLpix, and 40 hours for our method.


\noindent \textbf{Network Architectures and Optimizable Latents}. $Z_G$ and $Z_T$ both have a size of $128 \times 128 \times 16$. $\bigs{F_P}$ is based on Pix2PixHD \cite{pix2pixhd} architecture with Encoder blocks of [Conv2d, Batch- Norm, ReLU], ResNet \cite{He2016DeepRL} blocks, and Decoder blocks of [ReLU, ConvTranspose2d, BatchNorm]. $\bigs{F_P}$ has 3 Encoder and Decoder blocks, and 2 ResNet blocks. $\bigs{F_N}$ has 2 Decoder blocks. $\bigs{F_T}$ has $n$  ($n$ = 2 or 3 or 4) Encoder blocks, and the exact number depends on the downsampling factor of PD-NeRF such that the textural features and volumetric features have the same size as discussed at Sec. \ref{sec:tex_rendering}. $\bigs{F_R}$ has (${4-n}$) Encoder blocks, 4 Decoder blocks, and 5 ResNet blocks. For $\bigs{F_{\theta}}$, we use an 7-layer MLP with a skip connection from the input to the 4th layer as in DeepSDF \cite{park2019deepsdf}. From the 5th layer, the network branches out two heads, one to predict density with one fully-connected layer and the other one to predict color features with two fully-connected layers.

\begin{table*}[t]

	\begin{minipage}{\linewidth}
		\centering
		\begin{tabular}{lcccccc}
			\hline
			Models       & \multicolumn{1}{l}{LPIPS$\downarrow$} & \multicolumn{1}{c}{FID$\downarrow$} & \multicolumn{1}{c}{SSIM $\uparrow$} & \multicolumn{1}{c}{PSNR$\uparrow$} & \multicolumn{1}{c}{Time (s)} & \multicolumn{1}{c}{VR\_T(\%)} \\ \hline
			DNR          & 0.1023                    & 75.0152                 & 0.8310                   & 25.7303                  & 0.184                        & -                            \\
			SMPLpix      & 0.1002                    & 69.8119                 & 0.8350                   & 25.9295                  & 0.198                        & -                            \\
			ANR          & 0.1172                    & 78.5012                 & 0.8301                   & 26.0168                  & 0.224                        & -                            \\
			Neuray Body  & 0.2124                    & 155.8382                & 0.8328                   & 26.1718                  & 18.200                       & -                            \\
			Ours\_16(7)  & 0.0966                    & 64.8711                 & \textbf{0.8489}          & \textbf{26.4356}         & 0.292                        & 11.99                        \\
			Ours\_16(12) & 0.0911                    & 64.0892                 & 0.8417                   & 26.3272                  & 0.295                        & 12.88                        \\
			Ours\_16(20) & 0.0905                    & 62.8823                 & 0.8412                   & 26.3423         & 0.305                        & 15.41                        \\
			Ours\_8(12)  & 0.0901                    & 62.3330                 & 0.8415                   & 26.2165                  & 0.349                        & 26.36                        \\
			Ours\_4(20)  & \textbf{0.0861}           & \textbf{60.7884}        & 0.8461                   & 26.2465                  & 0.464                        & 44.61                        \\ \hline		
		\end{tabular}
		\caption{Performance, inference time of each methods. VR\_T(\%) indicates the percentages of the volume rendering time. Compared with Tab.~4 of the paper, the other two metrics SSIM and PSNR are included.}
		\label{tab:full_acc_time_supp}
	\end{minipage}
\end{table*}

\noindent \textbf{Geometry-guided Ray Marching}. The success of our method depends on the efficient and effective training of the pose-conditioned downsampled NeRF (PD-NeRF). First, instead of sampling rays in the whole space, we utilize a geometry-guided ray marching mesh as illustrated in Fig.~\ref{fig:dilation}. Specifically, we only sample query points along the corresponding rays near the SMPL \cite{smpl} mesh surface, which is determined by a dilated SMPL mesh. The SMPL mesh is dilated along the normal of each face with a radius of $d$, where $d$ is about 12cm for general clothes and 20cm for loose clothing like skirts for M1 dataset (see Fig. \ref{fig:skirt_geo}). We find the near and far points by querying the Z-buffer of the corresponding pixels after projecting the dilated SMPL mesh using Pytorch3D \cite{ravi2020pytorch3d}. In addition, we sample more points to the near region, which is expected to contain visible contents. The geometry-guided ray marching algorithm and UV conditioned architecture enable us to train a PD-NeRF with $45\times45$ resolution images and only 7 sampled point along each ray, as shown in Fig.~\ref{fig:geo45}. Though learned from low resolution images, the reconstructed geometry still preserves some pose-dependent features.

{\small
	\bibliographystyle{ieee_fullname}
	\bibliography{main}
}

\end{document}